\documentclass[twoside,11pt]{article}

\usepackage{jmlr2e}
\usepackage{amsmath}
\usepackage{amssymb}
\usepackage{graphicx,lipsum,afterpage}
\usepackage{color}
\usepackage{float}
\usepackage[caption = false]{subfig}

\usepackage{algorithm}
\usepackage[noend]{algpseudocode}

\ShortHeadings{Multivariate Bayesian Structural Time Series Model}{Jammalamadaka, Jinwen and Ning}

\begin{document}

\title{Multivariate Bayesian Structural Time Series Model}

\author{\name S. Rao Jammalamadaka \email rao@pstat.ucsb.edu \\
       \name Jinwen Qiu \email jqiu@pstat.ucsb.edu \\
       \name Ning Ning \email ning@pstat.ucsb.edu \\
       \addr Department of Statistics and Applied Probability\\
       University of California\\
       Santa Barbara, CA 93106, USA}

\editor{}

\maketitle

\begin{abstract}
	This paper deals with inference and prediction for multiple correlated time series, where one also has the choice of using a candidate pool of contemporaneous predictors for each target series. Starting with a structural model for time series, we use Bayesian tools for model fitting, prediction and feature selection, thus extending some recent works along these lines for the univariate case.
	The Bayesian paradigm in this multivariate setting helps the model avoid overfitting, as well as captures correlations among multiple target time series with various state components. The model provides needed flexibility in selecting a different set of components and available predictors for each target series. The cyclical component in the model can handle large variations in the short term, which may be caused by external shocks. Extensive simulations were run to investigate properties such as estimation accuracy and performance in forecasting. This was followed by an empirical study with one-step-ahead prediction on the max log return of a portfolio of stocks that involve four leading financial institutions. Both the simulation studies and the extensive empirical study confirm that this multivariate model outperforms three other benchmark models, viz. a model that treats each target series as independent, the autoregressive integrated moving average model with regression (ARIMAX), and the multivariate ARIMAX (MARIMAX) model.
\end{abstract}


\section{Introduction}

The analysis of ``Big Data" through the application of a new breed of analytical tools for manipulating and analyzing vast caches of data, is one of the cutting edge new areas. As a byproduct of the extensive use of the internet in collecting data on economic transactions, such data are growing exponentially every day. According to \cite{varian2014big} and the references therein, Google has $30$ trillion URLs and crawls over $20$ billion of those each day.
Conventional statistical and econometric techniques become increasingly inadequate to deal with such big data problems. For a good introduction to the new trends in data science, see \cite{Blei2017Science}.
Machine Learning as a field of computer science has strong ties to mathematical optimization and delivers methods, theory and applications, giving computers the ability to learn without being explicitly programmed (see a classical book, \cite{mohri2012foundations}).
Machine Learning indeed helps in developing high-performance computer tools, which often provide useful predictions in the presence of challenging computational needs. However, the result is one that we might call ``pure prediction" and is not necessarily based on substantive knowledge. Also, typical assumptions such as the data being independent and identically (or at least independently) distributed, are not satisfactory when dealing with time stamped data, which is driven by multiple ``predictors" or ``features".
We need to employ time series analysis for such series of data that are dependent,
such as macroeconomic indicators of the national economy, enterprise operational management, market forecasting, weather and hydrology prediction.

Our focus here is on new techniques that work well for feature selection problems in time series applications. \cite{scott2014predicting, scott2015bayesian} introduced and further explored the Bayesian Structural Time Series (BSTS) model, a technique that can be used for feature selection, time series forecasting, nowcasting, inferring causal relationships (regarding causality, see \cite{brodersen2015inferring} and \cite{peters2017elements}), among others.
One main ingredient of the BSTS model is that the time series aspect is handled through the Kalman filter (see \cite{harvey1990forecasting, durbin2002simple, petris2009dynamic}) while taking into account the trend, seasonality, regression, and other common time series factors. The second aspect is the ``spike and slab" variable selection, which was developed by \cite{george1997approaches} and \cite{madigan1994model}, by which the most important regression predictors are selected at each step. The third aspect is the Bayesian model averaging (see \cite{hoeting1999bayesian}), which combines the feature selection results and prediction calculation. All these three parts have natural Bayesian interpretations and tend to play well together so that the resulting BSTS model discovers not only correlations but also causations in the underlying data. Some excellent related literature includes, but is not limited to the following: \cite{dy2004feature, cortes1995support, guyon2003introduction, koo2007structured, bach2013hinge, keerthi2003asymptotic, nowozin2011structured, krishnapuram2005sparse, caron2006gps, csato2002sparse}.

In this paper, we extend the BSTS model to the multivariate target time series with various components, and label it the Multivariate Bayesian Structural Time Series (MBSTS) model.
For instance, the MBSTS model can be used to explicitly model the correlations between different stock returns in a portfolio through the covariance structure specified by $\Sigma_t$ (see Equation  \eqref{eq:1}). In this model, we allow  a cyclical component with a shock damping parameter to specially model the influence of a shock to the time series, in addition to a standard local linear trend component, a seasonal component, and a regression component. One motivation for this is provided by the 2007--2008 financial crisis to the stock market. In examples with simulated data, the properties of our model such as estimation and prediction accuracy is investigated. As an illustration, through an empirical case study, we predict the max log returns over $5$ consecutive business days of a stock portfolio with $4$ stocks: Bank of America (BOA), Capital One Financial Corporation (COF), J.P. Morgan (JPM) and Wells Fargo (WFC), using domestic Google trends and $8$ stock technical indicators as predictors.

Extensive analysis on both simulated data and real stock market data verifies that the MBSTS model gives much better prediction accuracy compared to the univariate BSTS model, the autoregressive integrated moving average with regression (ARIMAX) model, and the multivariate ARIMAX (MARIMAX) model. Some of the reasons for this can be seen in the following: the MBSTS model is strong in forecasting since it incorporates information of different components in the target time series, rather than merely historical values of the same component; the Bayesian paradigm and the MCMC algorithm can perform variable selection at the same time during model training and thus prevent overfitting, even if some spurious predictors are added into the candidate pool; the MBSTS model benefits from taking correlations among multiple target time series into account, which helps boost the forecasting power and is a significant improvement over the univariate BSTS model.

The rest of the paper is organized as follows. In Section 2, we build the basic model framework. Extensive simulations are carried out in Section 3 to examine how the model performs under various conditions. In Section 4, an empirical study on the stock portfolio is done to show how well our model performs with real-world data. Section 5 concludes with some final remarks.

\section{The MBSTS Model}
In this section, we introduce the MBSTS model including model structure, state components, prior elicitation and posterior inference. Then we describe the algorithm for training the model and performing forecasts. In the sequel, the symbol $``\sim "$ and the superscript $``(i)"$ will denote a column vector and the $i$-th component of a vector respectively, such as a $m\times1$ vector $\tilde{y}_t=[y_t^{(1)},\cdots,y_t^{(m)}]^T$.

\subsection{Structural Time Series}
Structural time series models belong to state space models for time series data given by the following set of equations:
\begin{equation} \label{eq:1}
	\tilde{y}_t=Z_t^T\alpha_t+\tilde{\epsilon}_t,  \ \ \ \ \  \ \tilde{\epsilon}_t\sim N_m(0,\Sigma_t),
\end{equation}
\begin{equation} \label{eq:2}
	\alpha_{t+1}=T_t\alpha_t+R_t\eta_t,  \ \ \ \ \  \eta_t\sim N_q(0,Q_t),
\end{equation}
\begin{equation} \label{eq:3}
	\alpha_0\sim N_d(\mu_0,\Sigma_0).
\end{equation}
Equation \eqref{eq:1} is called the observation equation, as it links the $m\times1$ vector $\tilde{y}_t$ of observations at time t with a $d \times 1$ vector $\alpha_t$ denoting the unobserved latent states, where $d$ is the total number of latent states for all entries in $\tilde{y}_t$.
Equation \eqref{eq:2} is called the transition equation because it defines how the latent states evolve over time. The model matrices $Z_t$, $T_t$, and $R_t$ typically contain unknown parameters and known values which are often set as 0 and 1.
$Z_t$ is a $d\times m$ output matrix, $T_t$ is a $d\times d$ transition matrix, and $R_t$ is a $d\times q$ control matrix.
The $m\times 1$ vector $\tilde{\epsilon}_t$ denotes observation errors with a $m\times m$ variance-covariance matrix $\Sigma_t$, and $\eta_t$  is a q-dimensional system error with a $q\times q$ state diffusion matrix $Q_t$, where $q\leq d$.
Note that any linear dependencies in the state vector can be moved from $Q_t$ to $R_t$, hence $Q_t$ can be set as a full rank variance matrix. 

Structural time series models constructed in terms of components have a direct interpretation. For example, one may consider the classical decomposition in which a series can be seen as the sum of trend, season, cycle and regression components. 
In general, the model in state space form can be written as:
\begin{equation} \label{eq:st}
	\tilde{y}_t=\tilde{\mu}_t+\tilde{\tau}_t+\tilde{\omega}_t+\tilde{\xi}_t+\tilde{\epsilon}_t,     \ \ \ \ \  \ \tilde{\epsilon}_t\stackrel{iid}\sim \  N_m(0,\Sigma_\epsilon),\ \ \ \ t=1,2\,\dots,n,
\end{equation}
where  $\tilde{y}_t,\ \tilde{\mu}_t,\ \tilde{\tau}_t,\ \tilde{\omega}_t,\ \tilde{\xi}_t \ $
and $\ \tilde{\epsilon}_t$ are m-dimension vectors, representing target time series, linear trend component, seasonal component, cyclical component, regression component and observation error terms respectively. Based on the  state space form, $\alpha_t$ is the collection of these components, namely  $\alpha_t=[\tilde{\mu}^T_t,\ \tilde{\tau}^T_t,\ \tilde{\omega}^T_t,\  \tilde{\xi}^T_t]^T$. Here $\Sigma_\epsilon$ is a $m\times m$ matrix, positive definite and is assumed to be constant over time for simplicity.
Structural time series models allow us to examine the time series and flexibly select suitable components for trend, seasonality, and either static or dynamic regression. In the current model, all state components are assembled independently, with each component yielding an additive contribution to $\tilde{y}_t$. The flexibility of the model allows us to include different model components
for each target series.

\subsection{Components of State}
The first component is a local linear trend. The specification of a time series model for the trend component varies according to the features displayed by the series under investigation and any prior knowledge. The most elementary structural model deals with
a series whose underlying level changes over time. Moreover, it also sometimes displays a steady upward or downward movement, suggesting to incorporate a slope or a drift into the model for the trend. The resulting model, a generalization of the local linear trend model where the slope exhibits stationarity instead of obeying a random walk, is expressed in the form as:
\begin{equation} \label{eq:trend}
	\tilde{\mu}_{t+1}=\tilde{\mu}_t+\tilde{\delta}_t+\tilde{u}_t, \ \ \  \ \ \tilde{u}_t\stackrel{iid}\sim \  N_m(0,\Sigma_\mu),
\end{equation}
\begin{equation} \label{eq:slope}
	\tilde{\delta}_{t+1}=\tilde{D}+\tilde{\rho}(\tilde{\delta}_t-\tilde{D})+\tilde{v}_t, \ \ \  \ \  \tilde{v}_t\stackrel{iid}\sim \  N_m(0,\Sigma_\delta),
\end{equation}
where $\tilde{\delta}_t$ and $ \tilde{D}$ are $m$-dimension vectors. $\tilde{\delta}_t$ is the expected increase in $\tilde{\mu}_t$  between times $t$ and $t+1$, so it can be thought as the slope at time $t$ and $\tilde{D}$ is the long-term slope. The parameter $\tilde{\rho}$
is a $m\times m$ diagonal matrix, whose diagonal entries $0\le \rho_{ii}\le1$ for $i=1,2,\cdots,m$, represent the learning rates at which the local trend is updated for $\{y_t^{(i)}\}_{i=1,2,\cdots,m}$. Thus, the model balances short-term information with long-term information.  When $\rho_{ii}=1$, the corresponding slope
becomes a random walk.

The second component is the one that captures seasonality. One frequently used model in the time domain is:
\begin{equation} \label{eq:seasonal}
	\tau_{t+1}^{(i)}=-\sum_{k=0}^{S_i-2}\tau^{(i)}_{t-k}+w^{(i)}_t,  \ \ \  \ \ \tilde{w}_t=[w^{(1)}_t,\cdots,w^{(m)}_t]^T\stackrel{iid}\sim  \ N_m(0,\Sigma_\tau),
\end{equation}
where $S_i$ represents the number of seasons for $y^{(i)}$ and a $m$-dimension vector $\tilde{\tau}_t$
denotes their joint contribution to the observed target time series $\tilde{y}_t$.
When we add a seasonal component, $S_i$ seasonal effects are set in the state space form for $y^{(i)}$. However, only one seasonal effect has error term based on equation \eqref{eq:seasonal} and other effects are represented by itself in a deterministic equation. More specifically, the part of the transition matrix $T_t$ representing the seasonal effects is an $(S_i-1)\times (S_i-1)$ matrix with $-1$ along the top row, $1$ along the subdiagonal and $0$ elsewhere.
In addition, the expectation of the summation of $S_i$ seasonal effects for $y^{(i)}$ is zero with variance equal to the $i$-th diagonal element of $\Sigma_\tau$.

For each target series $y^{(i)}$, the model allows for various seasonal components with different periods as shown in equation \eqref{eq:seasonal}.
For instance, we might include a seasonal component with $S_i=7$ to capture day-of-the-week effect for target series $y^{(i)}$, and $S_j=30$ indicating day-of-the-month effect for another target series $y^{(j)}$ when modeling daily data. The corresponding seasonal transition matrix in state space setting is a $6\times6$ matrix and a $29\times 29$ matrix with nonzero error variance for $y^{(i)}$ and $y^{(j)}$ respectively.

The third component is the one accounting for cyclical effects in the series. In economics, the term ``business cycle" broadly refers to recurrent, not exactly periodic, deviations around the long-term path of the series. A model with a cyclical component is capable of reproducing commonly acknowledged essential features, such as the presence of strong autocorrelation, recurrence and alternation of phases, dampening of fluctuations, and zero long run persistence. A stochastic trend model of a seasonally adjusted economic time series does not capture the
short-term movement of the series by itself. Including a serially correlated stationary component, the short-term movement could be captured, and this is the model incorporating cyclical effect (\cite{harvey2007trends}). The cycle component is postulated as:
\begin{equation} \label{eq:cycle}
	\begin{gathered}
		\tilde{\omega}_{t+1}=\tilde{\varrho}\widehat{\cos(\lambda)}\tilde{\omega}_{t}+\tilde{\varrho} \widehat{\sin(\lambda)}\tilde{\omega}_{t}^\star+\tilde{\kappa}_t, \ \ \ \ \ \  \tilde{\kappa}_t\stackrel{iid}\sim \  N_m(0,\Sigma_\omega),\\
		\tilde{\omega}_{t+1}^\star=-\tilde{\varrho}\widehat{\sin(\lambda)}\tilde{\omega}_{t}+\tilde{\varrho} \widehat{\cos(\lambda)}\tilde{\omega}_{t}^\star+\tilde{\kappa}_t^\star , \ \ \ \ \  \tilde{\kappa}_t^\star\stackrel{iid}\sim \  N_m(0,\Sigma_\omega),
	\end{gathered}
\end{equation}
where $\tilde{\varrho},\ \widehat{\sin(\lambda)},\ \widehat{\cos(\lambda)}$ are $m\times m$ diagonal matrices with diagonal entries equal to $\varrho_{ii}$ (a damping factor  for target series $y^{(i)}$ such that $0<\varrho_{ii}<1$), $\sin(\lambda_{ii})$ where $\lambda_{ii}=2\pi/q_i$ is the frequency with $q_i$ being a period such that $0<\lambda_{ii}<\pi$, and $\cos(\lambda_{ii})$ respectively. When $\lambda_{ii}$ is $0$ or $\pi$, the model degenerates to the AR(1) process. The damping factor should be strictly less than one for stationary purpose. When the damping factor is bigger than one, there will be no restriction for the cyclical movement, resulting in extending the amplitude of the cycle.

\begin{figure*}[h]
	\centering
	\includegraphics[width=1\linewidth]{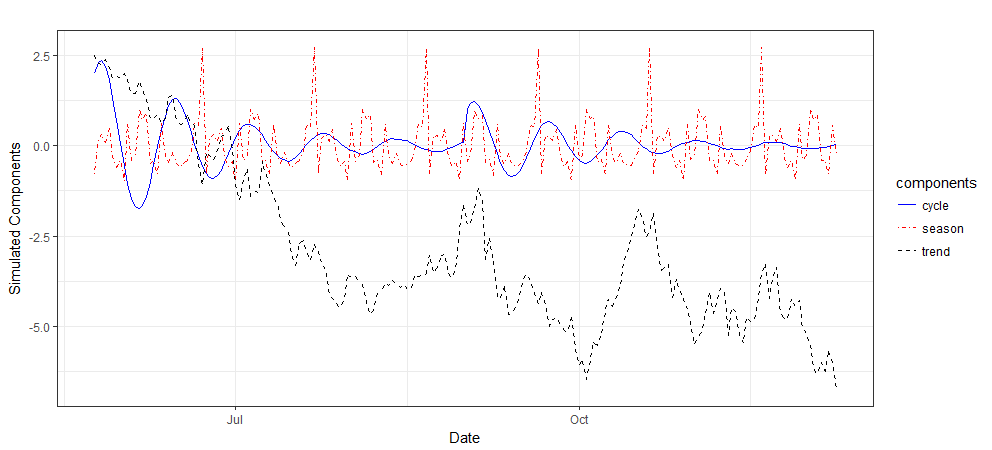}
	\caption{Simulated time series components include generalized linear trend, seasonality and cycle, generated by equations \eqref{eq:trend}, \eqref{eq:slope}, \eqref{eq:seasonal} and \eqref{eq:cycle} with $\tilde{\rho}=[0.6]$, $\tilde{D}=[0]$, $ \Sigma_\mu=[0.5^2]$, $\Sigma_\delta=[0.08^2]$, $ S=30$, $ \Sigma_\tau=[0.01^2]$, $ \lambda=\pi/10$, $\tilde{\varrho}=[0.97]$ and $\Sigma_\omega=[0.01^2]$, to show different contributions in explaining variations in target time series.}
	\label{simulatedstate}
\end{figure*}

These three time series components are illustrated in Figure \ref{simulatedstate}. The big difference between the cyclical component and the seasonal component is the damping factor. The amplitude of the cyclical component will decay as time goes by, which can be applied to target time series affected by external shocks. Here $\Sigma_\mu$, $\Sigma_\delta$, $\Sigma_\tau$ and $\Sigma_\omega$ are $m\times m$ variance-covariance matrices for error terms of different time series components, and for simplicity we assume they are diagonal.

The fourth component is the regression component with static coefficients written as follows:
\begin{equation} \label{eq:regression}
	\xi^{(i)}_t=\beta_i^Tx^{(i)}_t.
\end{equation}
Here $\tilde{\xi}_t=[\xi_{t}^{(1)},\cdots,\xi_{t}^{(m)}]^T$ is the collection of all elements in the regression component.
For target series $y^{(i)}$,  $x_t^{(i)}=[x_{t1}^{(i)},\dots,x_{tk_i}^{(i)}]^T$ is the pool of all available predictors at time t, and $\beta_i=[\beta_{i1},\dots,\beta_{ij},\dots,\beta_{ik_i}]^T$ represents corresponding static regression coefficients. All predictors are supposed to be contemporaneous with a known lag, which can be easily incorporated by shifting the corresponding predictors in time.

\subsection{Spike and Slab Regression}
In feature selection, a high degree of sparsity is expected, in the sense that coefficients of the vast majority of predictors are expected to be zero. A natural way to represent sparsity in the Bayesian paradigm is through the spike and slab coefficients. One advantage of working in a fully Bayesian setting is that we do not need to commit to a fixed set of predictors.

\subsubsection{Matrix Representation}
In order to assign appropriate prior distributions to parameters, we first combine  $\tilde{y}_t,\ \tilde{\mu}_t,\ \tilde{\tau}_t,\ \tilde{\omega}_t,$ $\tilde{\epsilon}_t$ into a $n\times m$ matrix as follows: $Y=[\tilde{y}_1,\dots,\tilde{y}_t,
\dots, \tilde{y}_n]^T$, $M=[\tilde{\mu}_1,\dots,\tilde{\mu}_t,
\dots, \tilde{\mu}_n]^T$,  $T=[\tilde{\tau}_1,\dots,\tilde{\tau}_t,
\dots, \tilde{\tau}_n]^T$, $W=[\tilde{\omega}_1,\dots,\tilde{\omega}_t,
\dots, \tilde{\omega}_n]^T$ and $E=[\tilde{\epsilon}_1,\dots,\tilde{\epsilon}_t,
\dots, \tilde{\epsilon}_n]^T$. Then the model can be written in a long matrix form as follows:

\begin{equation} \label{eq:matrixform}
	\tilde{Y}=\tilde{M}+\tilde{T}+\tilde{W}+X\beta+\tilde{E},
\end{equation}
where $\tilde{Y}=vec(Y)$, $\tilde{M}=vec(M)$, $\tilde{T}=vec(T)$, $ \tilde{W}=vec(W)$, $\tilde{E}=vec(E)$, and $X$, $\beta$ are written as:

\begin{equation} \label{eq:stack}
	X=\begin{bmatrix}
		X_1 & 0 & 0 & \dots  & 0 \\
		0 & X_2 & 0 & \dots  & 0 \\
		\vdots & \vdots & \vdots & \ddots & \vdots \\
		0 & 0 & 0 & \dots  & X_{m}
	\end{bmatrix},
	\quad\quad \beta= \begin{bmatrix}
		\beta_1  \\
		\beta_2  \\
		\vdots  \\
		\beta_m
	\end{bmatrix},
\end{equation}
where $X_i$ being a $n \times k_i$ matrix, representing all observations of  $k_i$ candidate
predictors for $y^{(i)}$, which is all observations of the $i$-th target series. The regression matrix $X$ is of dimension $(nm\times K)$ with
$K=\sum_{i=1}^{m} k_i$.
Moreover, $X_i$ and $X_j$ can be the same or only contain a portion of common predictors.
The regression coefficients for $y^{(i)}$ denoted as $\beta_i=[\beta_{i1},\dots,\beta_{ij},\dots,\beta_{ik_i}]^T$ is a $k_i$-dimension vector.
Reformulating the model in this way facilitates the mathematical derivation in selecting a different set of available predictors at each iteration for $y^{(i)}$.

\subsubsection{Prior distribution and elicitation}

We define $\gamma_{ij}=1$ if $\beta_{ij}\ne0$, and  $\gamma_{ij}=0$ if $\beta_{ij}=0$. Then $\gamma=[\gamma_1,\dots,\gamma_m]$ where $\gamma_i=[\gamma_{i1},\dots,\gamma_{ik_i}]$. Denote  $\beta_{\gamma}$ as the subset of elements of $\beta$ where $\beta_{ij}\ne 0$, and let $X_\gamma$ be the subset of columns of X where $\gamma_{ij}=1$.
The spike prior is written as:
\begin{equation} \label{eq:gama}
	\gamma\sim\prod_{i=1}^{m}\prod_{j=1}^{k_i}\pi_{ij}^{\gamma_{ij}}(1-\pi_{ij})^{1-\gamma_{ij}},\quad\quad i=1,\cdots, m,
\end{equation}
where $\pi_{ij}$ is the prior inclusion probability of the $j$-th predictor for the $i$-th target time series. Equation \eqref{eq:gama} is often further simplified by setting all the $\pi_{ij}$ for $j=1,2,\cdots,k_i$ as the same value $\pi_i$ for $y^{(i)}$ if prior information about effects of specific predictors on each target series are not available. With sufficient prior information available, assigning different subjectively determined values to $\pi_{ij}$ might provide more robust results without a great amount of computational burden. An easy way to elicit $\pi_i$ is to ask researchers for an ``expected model size", so that if one expects $q_i$ nonzero predictors for $y^{(i)}$, then $\pi_i=q_i/k_i$, where $k_i$ is the total number of candidate predictors for the $i$-th target series. Under some circumstances, $\pi_{ij}$ could be set as $0$ or $1$, for some specific predictors of $y^{(i)}$, forcing certain variables to be excluded or included. The spike prior can be specified by researchers in different distributional forms.

The natural conjugate prior for the multivariate model with the same set of predictors has the conjugate prior on $\beta$ depending on $\Sigma_\epsilon$. However, the multivariate extension with different set of predictors in each equation will destroy the conjugacy (\cite{rossi2012bayesian}). Conjugate priors such as the normal distribution and the inverse Wishart distribution can still be used in a nonconjugate context, since models can be conjugate conditional on some other parameters.
In order to obtain this conditional conjugate, we stack up the regression equations into one shown in equation \eqref{eq:stack}.
A simple slab prior specification is to make $\beta$ and $\Sigma_\epsilon$ prior independent (\cite{griffiths2003bayesian}):
\begin{equation} \label{eq:slab}
	\begin{gathered}
		p(\beta,\Sigma_\epsilon,\gamma)=p(\beta|\gamma)p(\Sigma_\epsilon|\gamma)p(\gamma),\\
		\beta|\gamma \sim N_K(b_\gamma,A_\gamma^{-1}),\\
		\Sigma_\epsilon|\gamma\sim IW(v_0,V_0),
	\end{gathered}
\end{equation}
where $b_\gamma$ is the vector of prior means 
and $A_\gamma=\kappa X^T_\gamma X_\gamma/n$ is the full-model prior information matrix, with $\kappa$ the number of observations worth of weight on the prior mean vector $b_\gamma$. If $X^T_\gamma X_\gamma$ is not positive definite due to perfect collinearity among predictors,   $A_\gamma=\kappa(\omega X_\gamma^TX_\gamma+(1-\omega)diag(X_\gamma^TX_\gamma))/n$ can be used instead to guarantee propriety. Given analysts' specification, $A_\gamma$ can be set in other forms. Here, $IW(v_0,V_0)$ is the inverse Wishart distribution with $v_0$ the number of degrees of freedom and $V_0$ a $m\times m$ scale matrix. Although these priors are not conjugate, they are conditionally conjugate.

Equation \eqref{eq:slab} is the so-called ``slab" because one can choose the prior parameters to make it only very weakly informative (close to flat), conditional on $\gamma$.  The vector $b_\gamma$ encodes our prior expectation about the value of each element of $\beta_\gamma$. In practice, one usually sets $b=0$. The values of $v_0$ and $V_0$ can be set by asking analysts for an expected $R^2$ form the regression, and a number of observations worth of weight $v_0$, which must be greater than the dimension of $\tilde{y}_t$ plus one. Then $V_0=(v_0-m-1)*(1-R^2)*\Sigma_y$, where $\Sigma_y$ is the variance-covariance matrix for multiple target time series $Y$.

Prior distributions of other variance-covariance matrices can be expressed as:
\begin{equation} \label{eq:15}
	\Sigma_u\sim IW(w_u,W_u), \quad\text{for } u\in \{\mu,\delta,\tau,\omega\}.
\end{equation}
By the assumption that all components are independent of each other,  the prior distributions in multivariate form can reduced to their univariate counterparts since the matrices are diagonal. In other words, each diagonal entry of these matrices follows inverse gamma distributions as introduced in BSTS.

\subsubsection{Posterior Inference}
By the law of total probability, the full likelihood function is given by
\begin{equation} \label{eq:liklihood}
	p(\tilde{Y}^\star,\beta,\Sigma_\epsilon,\gamma)=p(\tilde{Y}^\star|\beta,\Sigma_\epsilon,\gamma)\times p(\beta|\gamma)\times p(\Sigma_\epsilon|\gamma)\times p(\gamma),
\end{equation}
\begin{equation} \label{eq:y}
	p(\tilde{Y}^\star|\beta,\Sigma_\epsilon,\gamma)\propto |\Sigma_\epsilon|^{-n/2} \exp\left({-\frac{1}{2}(\tilde{Y}^\star-X_\gamma \beta_\gamma)^T(\Sigma_\epsilon^{-1}\otimes I_n)(\tilde{Y}^\star-X_\gamma \beta_\gamma)}\right),
\end{equation}
\begin{equation} \label{eq:beta}
	p(\beta|\gamma)\propto |A_\gamma|^{1/2}\exp\left(-\frac{1}{2}(\beta_\gamma-b_\gamma)^T A_\gamma(\beta_\gamma-b_\gamma)\right),
\end{equation}
\begin{equation} \label{eq:sigma}
	p(\Sigma_\epsilon|\gamma) \propto |\Sigma_\epsilon|^{-(v_0+m+1)/2}\exp\left(tr(-\frac{1}{2}V_0\Sigma_\epsilon^{-1})\right),
\end{equation}
where $\tilde{Y}^\star=\tilde{Y}-\tilde{M}-\tilde{T}-\tilde{W}$ is the multiple target time series $\tilde{Y}$ with time series components (trend, seasonality and cycle) subtracted out.
Conditional on $\Sigma_\epsilon$, one can introduce a normal prior, standardize the observations to remove correlation, and produce a posterior. However, we cannot find a convenient prior to integrate out $\Sigma_\epsilon$ from this conditional posterior. We tackle this issue by transforming equation $\tilde{Y}^\star=X\beta+\tilde{E}$ into a system with uncorrelated errors, using the square root of the variance-covariance matrix, $\Sigma_\epsilon=U^TU$. That is, if we multiply $((U^{-1})^T\otimes I_n)$ both sides of the equation, by the fact that $(U^{-1})^T\Sigma_\epsilon U^{-1}=I$, the transformed system has uncorrelated errors:
\begin{equation} \label{eq:39}
	\begin{gathered}
		\hat{Y}^\star=\hat{X}\beta+\hat{E},\ \ \  \hat{Y}^\star=((U^{-1})^T\otimes I_n)\tilde{Y}^\star,\ \ \    \hat{X}=((U^{-1})^T\otimes I_n)X,\\
		\mathrm{Var}(\hat{E})=\mathbb{E}((U^{-1})^T\otimes I_n)\tilde{E}\tilde{E}^T((U^{-1})^T\otimes I_n)]=I_m\otimes I_n.\\
	\end{gathered}
\end{equation}
Then the full conditional distribution of $\beta|\hat{Y}^\star,\Sigma_\epsilon,\gamma$ can be expressed as:
\begin{equation} \label{eq:40}
	p(\beta|\hat{Y}^\star,\Sigma_\epsilon,\gamma)\propto \exp\left(-\frac{1}{2}((\hat{Y}^\star-\hat{X}_\gamma \beta_\gamma)^T(\hat{Y}^\star-\hat{X}_\gamma \beta_\gamma)+(\beta_\gamma-b_\gamma)^T A_\gamma(\beta_\gamma-b_\gamma))\right).
\end{equation}
Let us combine the two terms in exponential:
\begin{equation} \label{eq:41}
	\begin{split}
		&(\hat{Y}^\star-\hat{X}_\gamma \beta_\gamma)^T(\hat{Y}^\star-\hat{X}_\gamma \beta_\gamma)+(\beta_\gamma-b_\gamma)^T A_\gamma(\beta_\gamma-b_\gamma)\\
		=&\beta^T_\gamma(\hat{X}_\gamma^T\hat{X}_\gamma+A_\gamma)\beta_\gamma-\beta^T_\gamma(\hat{X}_\gamma^T\hat{Y}^\star+A_\gamma b_\gamma)-(\hat{X}_\gamma^T\hat{Y}^\star+A_\gamma b_\gamma)^T\beta_\gamma+Const \\
		=&(\beta_\gamma-\tilde{\beta}_\gamma)^T(\hat{X}_\gamma^T\hat{X}_\gamma+A_\gamma)(\beta_\gamma-\tilde{\beta}_\gamma)+Const,
	\end{split}
\end{equation}
where $\tilde{\beta}_\gamma=(\hat{X}_\gamma^T\hat{X}_\gamma+A_\gamma)^{-1}(\hat{X}_\gamma^T\hat{Y}^\star+A_\gamma b_\gamma)$.
Then, a normal prior for $\beta_\gamma$ is conjugate with the conditional likelihood for the transformed system:
\begin{equation} \label{eq:posbeta}
	\beta|\hat{Y}^\star,\Sigma_\epsilon,\gamma \sim N_K(\tilde{\beta}_\gamma,(\hat{X}_\gamma^T\hat{X}_\gamma+A_\gamma)^{-1}).
\end{equation}
As $A_\gamma$ gets smaller, the prior becomes flatter. The mean  $\tilde{\beta}_\gamma$ can be recognized as the generalized least squares estimator.

The posterior of $\Sigma_\epsilon|\hat{Y}^\star,\beta,\gamma $ is in the inverted Wishart form. To see this, firstly recall that given $\beta_\gamma$ we can observe or compute the errors $\tilde{E}$. Then the problem becomes a standard inference problem of a variance-covariance matrix using a multivariate normal sample.
From equations \eqref{eq:liklihood}, \eqref{eq:y}, \eqref{eq:beta} and \eqref{eq:sigma}, we know that
\begin{equation} \label{eq:43}
	p(\Sigma_\epsilon|\tilde{Y}^\star,\beta,\gamma)\propto  |\Sigma_\epsilon|^{-(n+v_0+m+1)/2}\exp\left(-\frac{1}{2}\{\tilde{E}_\gamma^T(\Sigma_\epsilon^{-1}\otimes I_n)\tilde{E}_\gamma+tr(V_0\Sigma_\epsilon^{-1})\}\right),
\end{equation}
where $\tilde{E}_\gamma=\tilde{Y}^\star-X_\gamma \beta_\gamma$. The terms in the exponential part can be expressed in a trace form:
\begin{equation} \label{eq:44}
	\begin{split}
		\tilde{E}_\gamma^T(\Sigma_\epsilon^{-1}\otimes I_n)\tilde{E}_\gamma
		=vec(E_\gamma)^T(\Sigma_\epsilon^{-1}\otimes I_n)vec(E_\gamma)
		=tr(E_\gamma^TE_\gamma\Sigma_\epsilon^{-1}),
	\end{split}
\end{equation}
where $E_\gamma=Y^\star-X^\star_\gamma B_\gamma$, $Y^\star=Y-M-T-W$,  $X_\gamma^\star=[X_1,X_2,\dots,X_M]_\gamma$ is a $(n\times K)$ matrix, and $B_\gamma$ is a $(K\times m)$ matrix expressed as follows:
\begin{equation} \label{eq:45}
	B_\gamma=\begin{bmatrix}
		\beta_1 & 0 & 0 & \dots  & 0 \\
		0 & \beta_2 & 0 & \dots  & 0 \\
		\vdots & \vdots & \vdots & \ddots & \vdots \\
		0 & 0 & 0 & \dots  & \beta_{m}
	\end{bmatrix}_\gamma.
\end{equation}
Then the full conditional distribution of $\Sigma_\epsilon$ is inverted Wishart as follows:
\begin{equation} \label{eq:46}
	p(\Sigma_\epsilon|\tilde{Y}^\star,\beta,\gamma)\propto  |\Sigma_\epsilon|^{-(n+v_0+m+1)/2}\exp\left(-\frac{1}{2}\{tr[(E_\gamma^TE_\gamma+V_0)\Sigma_\epsilon^{-1}]\}\right),
\end{equation}
\begin{equation} \label{eq:47}
	\Sigma_\epsilon|\tilde{Y}^\star,\beta,\gamma\sim IW(v_0+n,E_\gamma^TE_\gamma+V_0).
\end{equation}
Note that, if we let the prior precision goes to zero, the posterior on $\Sigma_\epsilon$  would center over the sum of squared residuals matrices.

Since there is no conjugacy in this prior setting, we can not get an analytic solution of the marginal distribution of $\gamma$. However, the conditional distribution of $\gamma|\Sigma_\epsilon,\tilde{Y}^\star$ can be derived by the properties of conditional conjugacy. The joint probability density function $p(\Sigma_\epsilon,\tilde{Y}^\star,\gamma)$ can be obtained as follows:
\begin{equation} \label{eq:48}
	\begin{split}
		p(\Sigma_\epsilon,\tilde{Y}^\star,\gamma)= &\int_{-\infty}^{+\infty}p(\beta,\Sigma_\epsilon,\tilde{Y}^\star,\gamma)d\beta\\
		\propto&|\Sigma_\epsilon|^{-(v_0+m++n+1)/2}\exp\left(-\frac{1}{2}\{tr(V_0\Sigma_\epsilon^{-1})+(\hat{Y}^\star)^T\hat{Y}^\star\}\right)\\
		& \times\frac{|A_\gamma|^{1/2}p(\gamma)}{|\hat{X}_\gamma^T\hat{X}_\gamma+A_\gamma|^{1/2}}\exp\left(-\frac{1}{2}\{b_\gamma^TA_\gamma b_\gamma-Z_\gamma^T(\hat{X}_\gamma^T\hat{X}_\gamma+A_\gamma)^{-1} Z_\gamma\}\right),
	\end{split}
\end{equation}
where $Z_\gamma=(\hat{X}_\gamma^T\hat{Y}^\star+A_\gamma b_\gamma)$.
Then the conditional distribution of $\gamma|\Sigma_\epsilon,\tilde{Y}^\star$ can be expressed as:
\begin{equation} \label{eq:49}
	p(\gamma|\Sigma_\epsilon,\tilde{Y}^\star)=  C(\Sigma_\epsilon,\tilde{Y}^\star)\frac{|A_\gamma|^{1/2}p(\gamma)}{|\hat{X}_\gamma^T\hat{X}_\gamma+A_\gamma|^{1/2}}\exp\left(-\frac{1}{2}\{b_\gamma^TA_\gamma b_\gamma-Z_\gamma^T(\hat{X}_\gamma^T\hat{X}_\gamma+A_\gamma)^{-1} Z_\gamma\}\right),
\end{equation}
where $C(\Sigma_\epsilon,\tilde{Y}^\star)$ is a normalizing constant that only depends on $\Sigma_\epsilon$ and $\tilde{Y}^\star$. Note that,  matrices needed to be computed here are of low dimension, in the sense that equation \eqref{eq:49} places positive probabilities on coefficients being zero, leading to the sparsity of these matrices. In general, as a feature of the full posterior distribution, sparsity in this model enables equation \eqref{eq:49} to be evaluated in an inexpensive way.

Next we need to derive conditional posterior of $\Sigma_u$ for $u\in \{\mu,\delta,\tau,\omega\}$.
Given the draws of states, parameters drawn are straightforward for all state components except the static regression coefficients. All time series components that solely depend on their variance parameters would translate their draws back to the error terms and accumulate sums of squares. For the reason that inverse Wishart distribution is the conjugate prior of a multivariate normal distribution with known mean and variance-covariance, the posterior distribution is still inverse Wishart distributed
\begin{equation} \label{eq:30}
	\Sigma_u|u \sim IW(w_u+n,W_u+AA^T), \quad \text{for } u\in\{\mu,\delta,\tau,\omega\},
\end{equation}
where $A=[\tilde{A}_1,\dots,\tilde{A}_n]$ is a $m\times n$ matrix, representing a collection of residues of each time series component.

\subsection{Markov Chain Monte Carlo}
Markov chain Monte Carlo (MCMC) methods are a class of algorithms to sample from a probability distribution based on constructing a Markov chain that has the desired distribution as its equilibrium distribution. The state of the chain after a number of steps is then used as a sample from the desired distribution. The quality of the sample improves as an increasing function of the number of steps.

\subsubsection{Model Training}
Let $\theta=(\Sigma_\mu,\Sigma_\delta,\Sigma_\tau,\Sigma_\omega)$ denotes the set of state component parameters. The posterior distribution of the model can be simulated by a Markov chain Monte Carlo algorithm given in Algorithm \ref{algo:slide_generator}. Looping through the five steps yields a sequence of draws $\tilde{\psi}=(\alpha,\theta,\gamma,\Sigma_\epsilon, \beta)$ from a Markov chain with stationary distribution $p(\tilde{\psi}|Y)$, the posterior distribution of $\tilde{\psi}$ given $Y$.

\begin{algorithm}
	\caption{MBSTS Model Training}
	\label{algo:slide_generator}
	\begin{algorithmic}[1]
		\State Draw the latent state $\alpha=(\tilde{\mu},\tilde{\delta},\tilde{\tau},\tilde{\omega})$ from given model parameters and $\tilde{Y}$, namely $p(\alpha|\tilde{Y},\theta,\gamma,\Sigma_\epsilon,\beta)$, using the posterior simulation algorithm from \cite{durbin2002simple}.
		\State Draw time series state component parameters $\theta$ given $\alpha$, namely simulating
		$\theta \sim p(\theta|\tilde{Y},\alpha)$ based on equation \eqref{eq:30}.
		\State Loop over $i$ in an random order, draw each  $\gamma_i|\gamma_{-i},\tilde{Y},\alpha,\Sigma_\epsilon$, namely simulating $\gamma \sim p(\gamma|\tilde{Y}^\star,\Sigma_\epsilon)$  one by one based on equation \eqref{eq:49}, using the stochastic search variable selection (SSVS) algorithm from \cite{george1997approaches}.
		\State Draw $\beta$ given $\Sigma_\epsilon$, $\gamma$, $\alpha$ and $\tilde{Y}$, namely simulating $ \beta \sim p(\beta|\Sigma_\epsilon,\gamma, \tilde{Y}^\star)$  based on equation \eqref{eq:posbeta}.		
		\State Draw $\Sigma_\epsilon$ given $\gamma$, $\alpha$, $\beta$ and $\tilde{Y}$, namely simulating $\Sigma_\epsilon \sim p(\Sigma_\epsilon|\gamma,\tilde{Y}^\star,\beta)$  based on equation \eqref{eq:47}.
		
	\end{algorithmic}
\end{algorithm}

\subsubsection{Target Series Forecasting}
As typically in Bayesian data analysis, forecasts using our model are based on the posterior predictive distribution. Given draws of model parameters and latent states from their posterior distribution, we can draw samples from the posterior predictive distribution. Let $\hat{Y}$ represents the set of values to be forecast. The posterior predictive distribution of $\hat{Y}$ can be expressed as follows:

\begin{equation} \label{eq:50}
	p(\hat{Y}|Y)=\int p(\hat{Y}|\tilde{\psi})p(\tilde{\psi}|Y)d\tilde{\psi}
\end{equation}
where $\tilde{\psi}$ is the set of all the model parameters and latent states randomly drawn from $p(\tilde{\psi}|Y)$. We can draw samples of $\hat{Y}$ from $p(\hat{Y}|\tilde{\psi})$ by simply iterating equations \eqref{eq:trend}, \eqref{eq:slope}, \eqref{eq:seasonal}, \eqref{eq:cycle} and \eqref{eq:regression} to move forward from initial values of states $\alpha$ with initial values of parameters $\theta$, $\beta$ and $\Sigma_\epsilon$.
In the one-step-ahead forecast, we draw samples from the multivariate normal distribution with mean equal to $\tilde{\mu}_n+\tilde{\delta}_n+\sum_{k=0}^{S-2}\tilde{\tau}_{n-k}+\tilde{\varrho}\widehat{cos(\lambda)}\tilde{\omega}_{n}+\tilde{\varrho} \widehat{sin(\lambda)}\tilde{\omega}_{n}^\star+\beta^{(k)}x_{n+1}$ and variance equal to $\Sigma_\epsilon+\Sigma_{\mu}+\Sigma_{\tau}+\Sigma_\omega.$
Therefore, the samples drawn in this way have the same distribution as those simulated directly from the posterior predictive distribution.

Note that, the predictive probability density is not conditioned on parameter estimates, and inclusion or exclusion of predictors with static regression coefficients, all of which have been integrated out. Thus, through Bayesian model averaging, we commit neither to any particular set of covariates which helps avoid arbitrary selection, nor to point estimates of their coefficients which prevents overfitting. By the multivariate nature in our MBSTS model, the correlations among multiple target series are naturally taken into account, when sampling for prediction values of several target series.
The posterior predictive density in equation \eqref{eq:50}, is defined as a joint distribution over all predicted target series, rather than as a collection of univariate distributions, which enables us to properly forecast multiple target series as a whole instead of predicting them individually. This is crucial, especially when generating summary statistics, such as mean and variance-covariance from joint empirical distribution of forecast values.

\section{Application to Simulated Data}
In order to investigate the properties of our model, in this section, we analyze computer-generated data through a series of independent simulations. We generated multiple datasets with different time spans, local trends, number of regressors, dimensions of target series and correlations among two target series to analyze three aspects of generated data: accuracy in parameter estimation, ability to select the correct variables, and forecast performance of the model.

\subsection{Generated Data}
To check whether the estimation error and estimation standard deviation decrease as sample size increases, we built four different models in equation \eqref{eq:model1-4}, each of which generates two target time series data with different numbers of observations ($50$, $100$, $200$, $400$, $800$, $1600$, $3200$). These datasets are simulated using latent states and a static regression component with four explanatory variables, one of which has no effect on each target series with zero coefficient.
Specifically, each target series was generated with a different set of state components and explanatory variables, while the insignificant variable for each target series is not the same.

The latent states were generated using a local linear trend component with and without a global slope, a seasonality component with period equal to four, and/or a cyclical component with $\lambda=\pi/10$ for both target series. All initial values are drawn from normal distribution with a mean of zero. The detailed model description is presented as follows:

\begin{equation} \label{eq:model1-4}
	\begin{gathered}
		\tilde{y}_t=\tilde{\alpha}_t+B^T\tilde{x}_t+\tilde{\epsilon}_t\\
		Model\ 1:\ \tilde{y}_t=\tilde{\mu}_t+B^T\tilde{x}_t+\tilde{\epsilon}_t \ \ \ \ \  \tilde{\alpha}_t=\tilde{\mu}_t\\
		Model\ 2:\ \tilde{y}_t=\tilde{\mu}^\prime_t+B^T\tilde{x}_t+\tilde{\epsilon}_t \ \ \ \ \  \tilde{\alpha}_t=\tilde{\mu}_t^\prime\\
		Model\ 3:\ \tilde{y}_t=\tilde{\mu}_t^\prime+\tilde{\tau}_t+B^T\tilde{x}_t+\tilde{\epsilon}_t \ \ \ \ \  \tilde{\alpha}_t=\tilde{\mu}_t^\prime+\tilde{\tau_t}\\
		Model\ 4:\ \tilde{y}_t=\tilde{\mu}_t^\prime+\tilde{\tau}_t+\tilde{\omega}_t+B^T\tilde{x}_t+\tilde{\epsilon}_t \ \ \ \ \  \tilde{\alpha}_t=\tilde{\mu}_t^\prime+\tilde{\tau_t}+\tilde{\omega}_t
	\end{gathered}
\end{equation}

\begin{equation} \label{eq:modelregression}
	\begin{gathered}
		\tilde{\epsilon}_t\stackrel{iid}\sim N_2(0,\Sigma_\epsilon)\ \ \ \ \ \Sigma_\epsilon=\begin{bmatrix}
			1.1 & 0.7  \\
			0.7 & 0.9
		\end{bmatrix} \\
		B=\begin{bmatrix}
			2 & -1 & -0.5 & 0 \\
			-1.5 & 4 & 0 & 2.5
		\end{bmatrix}^T \ \ \ \ \
		\tilde{x}_t=[x_{t1},x_{t2},x_{t3},x_{t4}]^T\\
		x_{t1}\stackrel{iid}\sim N(5,5^2)\ \ \  x_{t2}\stackrel{iid}\sim Pois(10)\ \ \ x_{t3}\stackrel{iid}\sim B(1,0.5)\ \ \  x_{t4}\stackrel{iid}\sim N(-2,5^2)
	\end{gathered}
\end{equation}

\begin{equation} \label{eq:sim-mu}
	\begin{gathered}
		\tilde{\mu}_{t+1}=\begin{bmatrix}
			\mu_{1,t+1}  \\
			\mu_{2,t+1}
		\end{bmatrix}=\begin{bmatrix}
			\mu_{1,t}  \\
			\mu_{2,t}
		\end{bmatrix}+\begin{bmatrix}
			\delta_{1,t}  \\
			0
		\end{bmatrix}+\begin{bmatrix}
			u_{1,t}  \\
			u_{2,t}
		\end{bmatrix}\\
		\delta_{1,t}\stackrel{iid}\sim N(\delta_{1,t-1},0.08^2)\quad\quad
		\begin{bmatrix}
			u_{1,t}  \\
			u_{2,t}
		\end{bmatrix}\stackrel{iid}\sim N_2\Bigg(
		\begin{bmatrix}
			0  \\
			0
		\end{bmatrix},
		\begingroup 
		\setlength\arraycolsep{1.8pt}
		\begin{bmatrix}
			0.5^2&0\\
			0&1
		\end{bmatrix}
		\endgroup\Bigg)
	\end{gathered}
\end{equation}

\begin{equation} \label{eq:sim-mu2}
	\begin{gathered}
		\tilde{\mu}^\prime_{t+1}=\begin{bmatrix}
			\mu_{1,t+1}^\prime  \\
			\mu_{2,t+1}^\prime
		\end{bmatrix}=\begin{bmatrix}
			\mu_{1,t}^\prime  \\
			\mu_{2,t}^\prime
		\end{bmatrix}+\begin{bmatrix}
			\delta_{1,t}^\prime  \\
			\delta_{2,t}^\prime
		\end{bmatrix}+\begin{bmatrix}
			u_{1,t}  \\
			u_{2,t}
		\end{bmatrix} \\
		\begin{bmatrix}
			\delta_{1,t}^\prime  \\
			\delta_{2,t}^\prime
		\end{bmatrix}\stackrel{iid}\sim N_2\Bigg(
		\begin{bmatrix}
			0.6\delta_{1,t-1}^\prime+0.4*0.02  \\
			\delta_{2,t-1}^\prime
		\end{bmatrix},
		\begingroup 
		\setlength\arraycolsep{1.8pt}
		\begin{bmatrix}
			0.08^2&0\\
			0&0.16^2
		\end{bmatrix}
		\endgroup\Bigg)
	\end{gathered}
\end{equation}

\begin{equation} \label{eq:sim-seasonal}
	\begin{gathered}
		\tilde{\tau}_{t+1}=\begin{bmatrix}
			\tau_{1,t+1}  \\
			\tau_{2,t+1}
		\end{bmatrix}=\begin{bmatrix}
			-\sum_{k=0}^{2}\tau_{1,t-k}  \\
			0
		\end{bmatrix}+\begin{bmatrix}
			w_{1,t}  \\
			0
		\end{bmatrix}\quad \quad
		w_{1,t}\stackrel{iid}\sim N(0,0.01^2)
	\end{gathered}
\end{equation}

\begin{equation} \label{eq:sim-cycle}
	\begin{gathered}
		\tilde{\omega}_{t+1}=\begin{bmatrix}
			\omega_{1,t+1}  \\
			\omega_{2,t+1}
		\end{bmatrix}=\begin{bmatrix}
			0  \\
			0.5*\cos(\lambda_{22})\omega_{2,t}
		\end{bmatrix}+\begin{bmatrix}
			0  \\
			0.5*\sin(\lambda_{22})\omega_{2,t}^\star
		\end{bmatrix}+\begin{bmatrix}
			0  \\
			\kappa_{2,t}
		\end{bmatrix} \\
		\tilde{\omega}_{t+1}^\star=\begin{bmatrix}
			\omega_{1,t+1}^\star  \\
			\omega_{2,t+1}^\star
		\end{bmatrix}=\begin{bmatrix}
			0  \\
			-0.5*\sin(\lambda_{22})\omega_{2,t}
		\end{bmatrix}+\begin{bmatrix}
			0  \\
			0.5*\cos(\lambda_{22})\omega_{2,t}^\star
		\end{bmatrix}+\begin{bmatrix}
			0  \\
			\kappa_{2,t}^\star
		\end{bmatrix}\\
		\kappa_{2,t}\stackrel{iid}\sim N(0,0.01^2) \ \ \ \ \
		\kappa_{2,t}^\star\stackrel{iid}\sim N(0,0.01^2).
	\end{gathered}
\end{equation}

To check the model performance with more than two series, two more datasets were generated by Model $5$ and Model $6$ according to equations \eqref{eq:model5} and \eqref{eq:model6}, respectively, where for simplicity we consider latent states only include a generalized local linear trend with and without a global slope. The specific settings are given below:

\begin{equation} \label{eq:model5}
	\begin{gathered}
		Model\ 5:\ \tilde{y}_t=\tilde{\mu}^{\prime\prime}_t+B^T\tilde{x}_t+\tilde{\epsilon}_t\ \ \ \ \ \ \tilde{\epsilon}_t\stackrel{iid}\sim N_3(0,\Sigma_\epsilon)\\
		B=\begin{bmatrix}
			2 & -1 & -0.5 & 0 \\
			-1.5 & 4 & 0 & 2.5 \\
			3 & 0 & 3.5 & -2 
		\end{bmatrix}^T \ \ \ \ \Sigma_\epsilon=\begin{bmatrix}
			1.1 & 0.7 & 0.7  \\
			0.7 & 0.9 & 0.7 \\
			0.7 & 0.7 & 1.0
		\end{bmatrix}\\
		\tilde{\mu}^{\prime\prime}_{t+1}=\begin{bmatrix}
			\mu_{1,t+1}^{\prime\prime}  \\
			\mu_{2,t+1}^{\prime\prime} \\
			\mu_{3,t+1}^{\prime\prime}
		\end{bmatrix}=\begin{bmatrix}
			\mu_{1,t}^{\prime\prime}  \\
			\mu_{2,t}^{\prime\prime} \\
			\mu_{3,t}^{\prime\prime}
		\end{bmatrix}+\begin{bmatrix}
			\delta_{1,t}^{\prime\prime}  \\
			\delta_{2,t}^{\prime\prime} \\
			\delta_{3,t}^{\prime\prime}
		\end{bmatrix}+\begin{bmatrix}
			u_{1,t}  \\
			u_{2,t} \\
			u_{3,t}
		\end{bmatrix}\\
		\begin{bmatrix}
			\delta_{1,t}^{\prime\prime}  \\
			\delta_{2,t}^{\prime\prime} \\
			\delta_{3,t}^{\prime\prime} 
		\end{bmatrix}\stackrel{iid}\sim N_3\Bigg(
		\begin{bmatrix}
			0.6\delta_{1,t-1}^{\prime\prime}+0.4*0.02  \\
			\delta_{2,t-1}^{\prime\prime} \\
			0.3\delta_{3,t-1}^{\prime\prime}+0.7*0.01
		\end{bmatrix},
		\begingroup 
		\setlength\arraycolsep{1.8pt}
		\begin{bmatrix}
			0.08^2&0 & 0\\
			0&0.16^2 & 0\\
			0& 0 & 0.12^2
		\end{bmatrix}
		\endgroup\Bigg)\\
		\begin{bmatrix}
			u_{1,t}  \\
			u_{2,t} \\
			u_{3,t}
		\end{bmatrix}\stackrel{iid}\sim N_3\Bigg(
		\begin{bmatrix}
			0  \\
			0 \\
			0
		\end{bmatrix},
		\begingroup 
		\setlength\arraycolsep{1.8pt}
		\begin{bmatrix}
			0.5^2&0 & 0\\
			0&1 & 0\\
			0 & 0 & 0.7^2
		\end{bmatrix}
		\endgroup\Bigg).
	\end{gathered}
\end{equation}

\begin{equation} \label{eq:model6}
	\begin{gathered}
		Model\ 6:\ \tilde{y}_t=\tilde{\mu}_t^{\prime\prime\prime}+B^T\tilde{x}_t+\tilde{\epsilon}_t\ \ \ \ \ \ \tilde{\epsilon}_t\stackrel{iid}\sim N_4(0,\Sigma_\epsilon)\\
		B=\begin{bmatrix}
			2 & -1 & -0.5 & 0 \\
			-1.5 & 4 & 0 & 2.5 \\
			3 & 0 & 3.5 & -2 \\
			0 & 1 & 1.5 & -0.5 
		\end{bmatrix}^T \ \ \ \ \Sigma_\epsilon=\begin{bmatrix}
			1.1 & 0.7 & 0.7 & 0.7\\
			0.7 & 0.9 & 0.7 & 0.7\\
			0.7 & 0.7 & 1.0 & 0.7\\
			0.7 & 0.7 & 0.7 & 1.2 \\
		\end{bmatrix}\\
		\tilde{\mu}^{\prime\prime\prime}_{t+1}=\begin{bmatrix}
			\mu_{1,t+1}^{\prime\prime\prime}  \\
			\mu_{2,t+1}^{\prime\prime\prime} \\
			\mu_{3,t+1}^{\prime\prime\prime} \\
			\mu_{4,t+1}^{\prime\prime\prime}
		\end{bmatrix}=\begin{bmatrix}
			\mu_{1,t}^{\prime\prime\prime} \\
			\mu_{2,t}^{\prime\prime\prime} \\
			\mu_{3,t}^{\prime\prime\prime} \\
			\mu_{4,t}^{\prime\prime\prime}
		\end{bmatrix}+\begin{bmatrix}
			\delta_{1,t}^{\prime\prime\prime}  \\
			\delta_{2,t}^{\prime\prime\prime} \\
			\delta_{3,t}^{\prime\prime\prime} \\
			\delta_{4,t}^{\prime\prime\prime}
		\end{bmatrix}+\begin{bmatrix}
			u_{1,t}  \\
			u_{2,t} \\
			u_{3,t} \\
			u_{4,t}
		\end{bmatrix} \\
		\begin{bmatrix}
			\delta_{1,t}^{\prime\prime\prime}  \\
			\delta_{2,t}^{\prime\prime\prime} \\
			\delta_{3,t}^{\prime\prime\prime} \\
			\delta_{4,t}^{\prime\prime\prime} 
		\end{bmatrix}\stackrel{iid}\sim N_4\Bigg(
		\begin{bmatrix}
			0.6\delta_{1,t-1}^{\prime\prime\prime}+0.4*0.02  \\
			\delta_{2,t-1}^{\prime\prime\prime} \\
			0.3\delta_{3,t-1}^{\prime\prime\prime}+0.7*0.01 \\
			0.5\delta_{4,t-1}^{\prime\prime\prime}
		\end{bmatrix},
		\begingroup 
		\setlength\arraycolsep{1.8pt}
		\begin{bmatrix}
			0.08^2&0 & 0 & 0\\
			0&0.16^2 & 0 & 0\\
			0& 0 & 0.12^2 & 0\\
			0& 0 & 0 & 0.10^2
		\end{bmatrix}
		\endgroup\Bigg)\\
		\begin{bmatrix}
			u_{1,t}  \\
			u_{2,t} \\
			u_{3,t} \\
			u_{4,t}
		\end{bmatrix}\stackrel{iid}\sim N_4\Bigg(
		\begin{bmatrix}
			0  \\
			0 \\
			0 \\
			0
		\end{bmatrix},
		\begingroup 
		\setlength\arraycolsep{1.8pt}
		\begin{bmatrix}
			0.5^2&0 & 0 & 0\\
			0&1 & 0 & 0\\
			0 & 0 & 0.7^2 & 0\\
			0 & 0 & 0 & 0.6^2
		\end{bmatrix}
		\endgroup\Bigg).
	\end{gathered}
\end{equation}

Model $7$ was used to generate data to examine the accuracy in Bayesian point and interval estimations and covariates inclusion probabilities. The model is described as follows:
\begin{equation} \label{eq:model7}
	\begin{gathered}
		Model\ 7:\
		\tilde{y}_t=\tilde{\mu}_t^\prime+\tilde{\tau}_t+\tilde{\omega}_t+diag(B^T\tilde{x}_t)+\tilde{\epsilon}_t \ \ \ \  \tilde{\epsilon}_t\stackrel{iid}\sim N_2(0,\Sigma_\epsilon)\\
		B=\begin{bmatrix}
			2 & -1 & -0.5 & 0 & 1.5 & -2 & 0 & 3.5\\
			-1.5 & 4 & 0 & 2.5  & -1 & 0 & -3 & 0.5
		\end{bmatrix}^T \\
		\tilde{x}_t=\begin{bmatrix}
			x_{t1} & x_{t2} & x_{t3}&  x_{t4}& x_{t5} & x_{t6}& x_{t7} & x_{t8}^\star\\
			x_{t1} & x_{t2}^\star & x_{t3}&  x_{t4}& x_{t5} & x_{t6}& x_{t7} & x_{t8}
		\end{bmatrix}^T\\
		x_{t5}\stackrel{iid}\sim N(-5,5^2)\ \ \  x_{t6}\stackrel{iid}\sim Pois(15)\ \ \  x_{t7}\stackrel{iid}\sim Pois(20)\ \ \ x_{t8}\stackrel{iid}\sim N(0,10^2),
	\end{gathered}
\end{equation}
where  $x_2^\star$, $x_5^\star$ and $x_8^\star$ are variables whose values were obtained by rearranging a partial portion of data values for $x_2$, $x_5$ and $x_8$, and the $diag()$ operator extracts diagonal entries in the matrix to form a column vector. In Model $7$, the first target series was generated by ($x_1$, $x_2$, $x_3$, $x_4$, $x_5$, $x_6$, $x_7$, $x_8^\star$) and the second target series was generated by ($x_1$, $x_2^\star$, $x_3$, $x_4$, $x_5$, $x_6$, $x_7$, $x_8$). Therefore, when explanatory variables ($x_1$, $x_2^\star$, $x_3$, $x_4$, $x_5^\star$, $x_6$, $x_7$, $x_8^\star$) are used for model training, regression coefficients of $x_2^\star$ (resp. $x_5^\star$) for the first target series generation are expected not to reflect the true linear relationship between $y^{(1)}$ and $x_2$ (resp.  $x_5$). Similarly, regression coefficients of $x_5^\star$ (resp. $x_8^\star$) for the second target series generation are expected not to reflect the true linear relationship between $y^{(2)}$ and $x_5$ (resp.  $x_8$).
In sum, each distinct target series has a unique pattern generated by a particular set of explanatory variables and state components (the first target series affected by seasonality, not cyclical effect; the second target series affected by cyclical effect, not seasonality).
Then we apply the MBSTS model on generated datasets to study its different properties.

\subsection{Estimation and Model Selection Accuracy}
\begin{figure*}[h]
	\subfloat[Model 1]{\includegraphics[width = 3in]{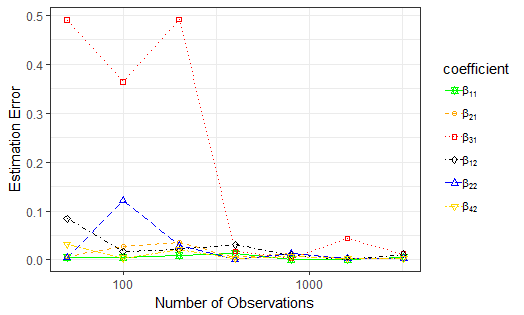}}
	\subfloat[Model 2]{\includegraphics[width = 3in]{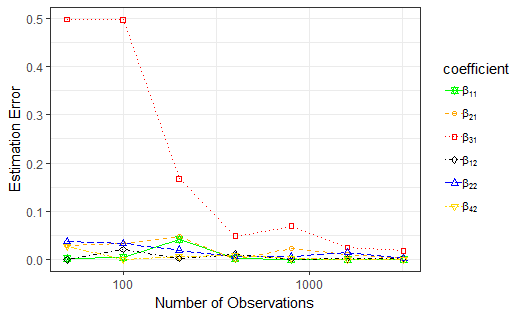}}\\
	\subfloat[Model 3]{\includegraphics[width = 3in]{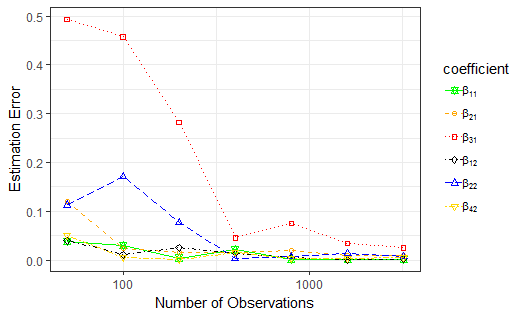}}
	\subfloat[Model 4]{\includegraphics[width = 3in]{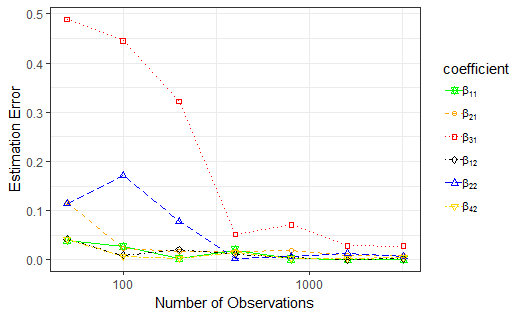}}
	\caption{Estimation error for regression coefficients with different sample size. (a), (b), (c) and (d) display results using generated datasets by four different models in equation \eqref{eq:model1-4}.}
	\label{fig:esterr}
\end{figure*}

\begin{figure*}[h]
	\subfloat[Model 1]{\includegraphics[width = 3in]{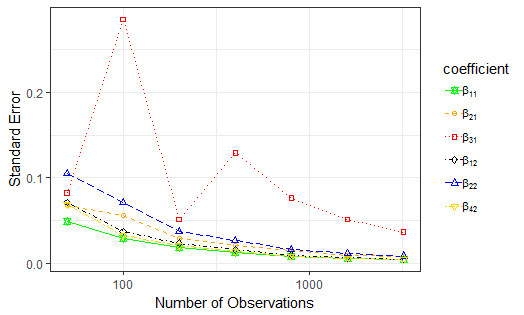}}
	\subfloat[Model 2]{\includegraphics[width = 3in]{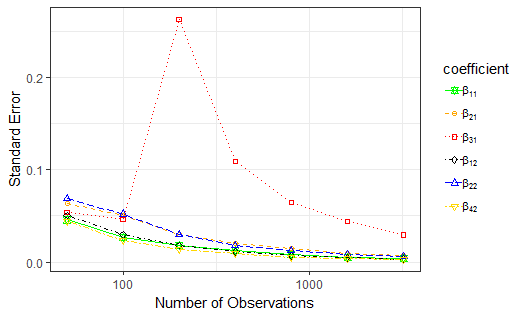}}\\
	\subfloat[Model 3]{\includegraphics[width = 3in]{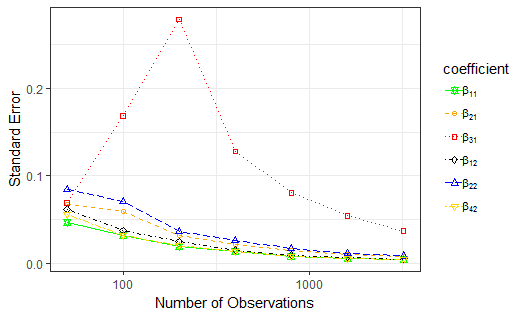}}
	\subfloat[Model 4]{\includegraphics[width = 3in]{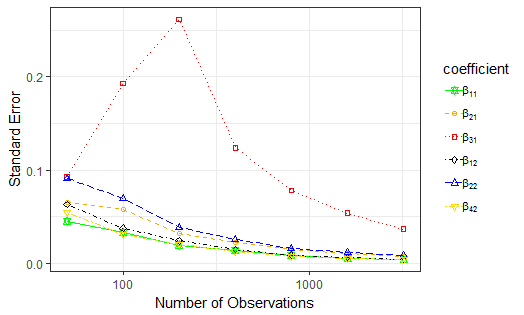}}
	\caption{Standard error for regression coefficients with different sample size. Here, standard error is the empirical standard deviation of draws from equation (22). (a), (b), (c)  and (d) display results using generated datasets by four different models in equation \eqref{eq:model1-4}.}
	\label{fig:std}
\end{figure*}

\begin{figure*}[h]
	\subfloat[Boxplot with $90\%$ Credible Interval]{\includegraphics[width = 3in]{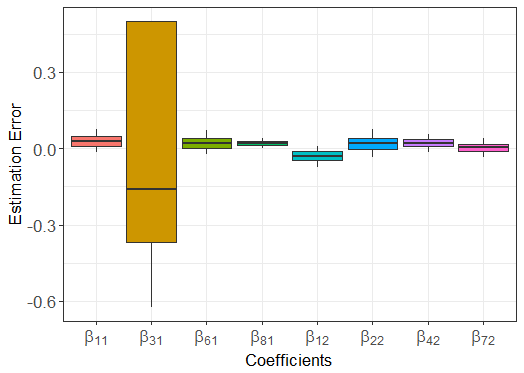}}
	\subfloat[Inclusion Probability]{\includegraphics[width = 3in]{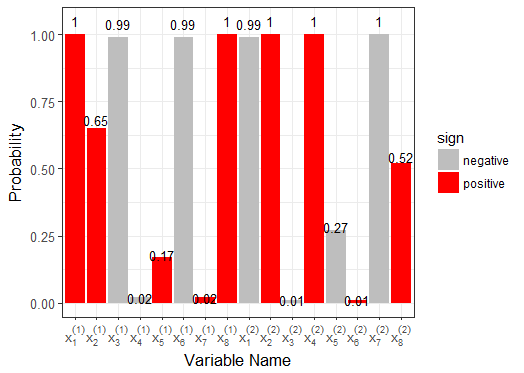}}
	\caption{Empirical posterior distribution of estimated coefficients and indicators. (a) Box plots of the difference between draws from equation (22) and true values of regression coefficients. The top and bottom correspond to the $95\%$ upper bound and $5\%$ low bound, respectively. (b) Bar plot of empirical inclusion probability illustrates the proportion of Monte Carlo draws with $\gamma_{ij}=1$. The red color shows positive estimated values of regression coefficients, while gray color displays negative values.}
	\label{Dis-inc}
\end{figure*}

From three perspectives, we explored properties of our model. More specifically, they include how the number of observations affects Bayesian estimation accuracy, how likely the $90\%$ credible interval contains the true values of coefficients, and how possible the model selects the most important explanatory variables and ignores variables that do not contribute as desired, with results given in Figures \ref{fig:esterr}, \ref{fig:std} and \ref{Dis-inc}, respectively.

With the advent of the ``big data" era, a huge amount of time series data are available to be analyzed from various sources. In the first analysis, we want to check whether a larger sample size improves the model performance in terms of Bayesian point estimation accuracy. After model training, we drew $2000$ samples for each coefficient to be estimated during MCMC iterations. 
To reduce the influence of initial values on posterior inferences, we discarded an initial portion of the Markov chain samples. Specifically based on trial and error, the first $200$ drawn samples were removed and the rest of them were used to build a sample posterior distribution for each parameter.

Based on the theory of Bayesian estimation, the sample mean from posterior distribution is considered to be the best point estimator for unknown parameters in terms of the mean squared error. We firstly consider the estimation error defined as the absolute value of difference between the true value and its Bayesian point estimate. The plots in Figure \ref{fig:esterr} illustrate how estimation errors of coefficients change as the sample size increases. The first target series was generated not using covariate $x_4$, while the second target series was generated not using covariate $x_3$, as shown in equation \eqref{eq:modelregression}. Those zero coefficients are not displayed in these line plots. Figure \ref{fig:esterr} shows that only the estimation error for coefficient $\beta_{31}$ goes down dramatically when sample size expands in these four cases. The remaining estimation errors stay almost the same regardless of different sample sizes, which implies that the number of observations significantly affect only the point estimation accuracy of coefficients for binary variables, not for numerical or ordinal variables. Even if only a small amount of data is available, our approach still performs well when binary or factor variables are not involved in the analysis. 

The sample standard error, defined as the posterior standard deviation of the regression coefficient, is used to illustrate the spread of the Bayesian estimator. 
To further explore other properties of the posterior distribution of draws, the standard errors were checked for each coefficient with different sample sizes. Figure \ref{fig:std} shows that all standard errors for covariates' coefficients except $\beta_{31}$ gradually decline with a larger amount of data. The standard error for coefficients $\beta_{31}$ peaks when the number of observations is $100$ for Model $1$ or $200$ for models $2$, $3$ and $4$, and then begin to drop very sharply. In general, a larger sample size helps shrink the standard errors of all coefficients, especially for binary or factor covariates' coefficients, as one would expect. In other words, collecting more data allows us to shrink the dispersion of the posterior empirical distribution from Monte Carlo draws, and hence build a narrower credible interval.

In the second analysis, we assess the coverage properties of the posterior credible intervals based on the empirical posterior distribution of each covariate's coefficients. In other words, the $90\%$ credible interval contains the ground truth in $90\%$ of the simulations. In Model $7$ (equation \eqref{eq:model7}), $x_8^\star$ instead of $x_8$ was used to generate the first target series $y^{(1)}$, and $x_2^\star$ instead of $x_2$ was used to generate the second target series $y^{(2)}$.
Therefore, when the explanatory variables ($x_1$, $x_2^\star$, $x_3$, $x_4$, $x_5^\star$, $x_6$, $x_7$, $x_8^\star$) are used for model training, the resulting coefficients $\beta_{21}$ and $\beta_{51}$ for $y^{(1)}$ as well as $\beta_{52}$ and $\beta_{82}$ for $y^{(2)}$, cannot reveal a true linear relationship. The box plot in Figure \ref{Dis-inc} displays the empirical posterior distribution of estimated coefficients for significant explanatory variables whose values were not randomly shuffled, and indicates that the true values of all coefficients are within $90\%$ credible intervals. In addition, we can see that the $90\%$ credible interval of binary covariates' coefficients is much wider than others, due to their larger standard errors.


In the third analysis, one important property of our model is to reduce data dimension by variable selection in model training. In Figure \ref{Dis-inc}, the bar plot of empirical inclusion probabilities based on the proportion of MCMC draws shows a clear picture of which variables are used to generate data and which are the ones with shuffled values.
For the first target series $y^{(1)}$, the empirical inclusion probabilities of covariates $x_1$, $x_3$, $x_6$ and $x_8$ as one or close to one indicate that they were all, or almost all, selected during MCMC iterations,
which is exactly how the dataset was generated; the covariates $x_4$ and $x_7$ with zero coefficients indicate that they are rarely selected during MCMC iterations. Some covariates with partially shuffled values, such as $x_2$ and $x_5$, are more likely be selected than those with no effect on this target series, but they are not so important as $x_1$, $x_3$, $x_6$ and $x_8$. Similar striking results were achieved for the second target series. Moreover, we can see that as expected, the inclusion probability of $x_5^\star$ is just $0.17$ (resp. $0.27$) for the first (resp. second) target series. In a word, our MBSTS model is good at variables selection, even if the variation of each target series is explained by a different set of explanatory variables.

It is worth emphasizing that our model performs very well in terms of estimation accuracy and variables selection ability, even if each target series has a different set of latent states and explanatory variables from others. However, all preceding results depend on the assumption that the model structure remains intact throughout the modeling period. In other words, even though the model is built on the idea of multiple non-stationary components such as a time-varying local trend, seasonal effect, and potentially dynamic regression coefficients, the structure itself remains unchanged. If the model structure does change over time (e.g. local trend disappears or the static regression coefficients become dynamic), the estimation accuracy may suffer. Therefore, a preliminary data exploration and acquiring a background knowledge about the dataset before applying our model is suggested, although it has the strength in allowing users to adjust the model components flexibly for each target series.

\subsection{Model Performance Comparison}
The generated datasets were split into a certain period of training data and a subsequent period of testing set. The standard approach would use the training data to develop the model that would then be applied to obtain predictions for the testing period. We use a growing window approach, which simply adds one new observation in the test set to the existing training set, obtaining a new model with fresher data and then constantly forecasting a new value in the test set.


\begin{figure*}[h]
	\subfloat[Model 1]{\includegraphics[width = 3in]{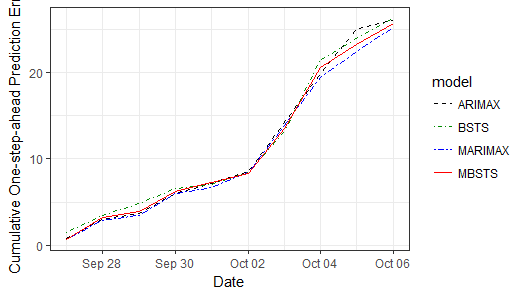}}
	\subfloat[Model 2]{\includegraphics[width = 3in]{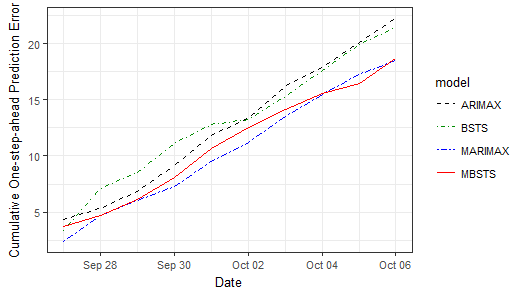}}\\
	\subfloat[Model 3]{\includegraphics[width = 3in]{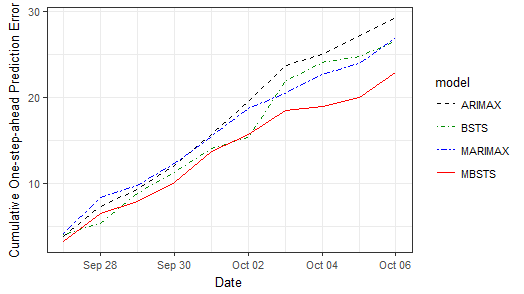}}
	\subfloat[Model 4]{\includegraphics[width = 3in]{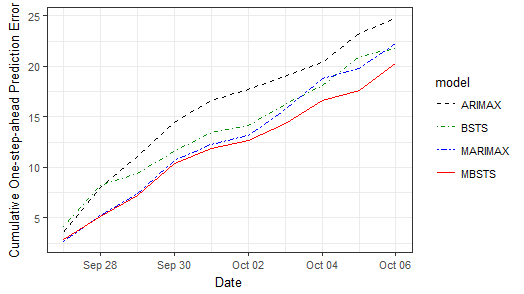}}\\
	\subfloat[Model 5]{\includegraphics[width = 3in]{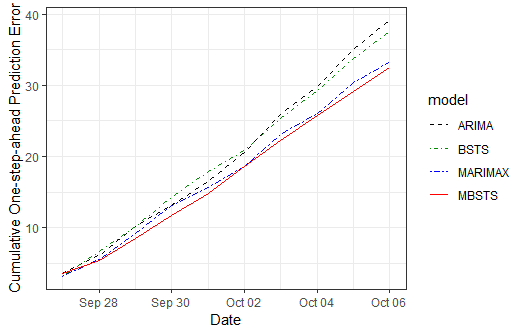}}
	\subfloat[Model 6]{\includegraphics[width = 3in]{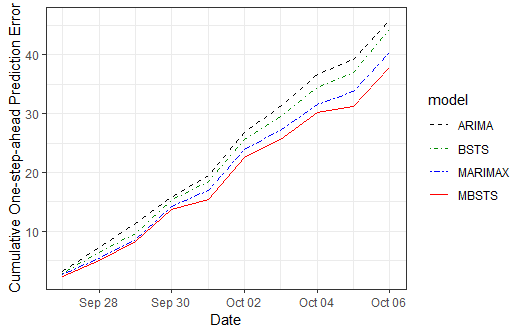}}
	\caption{Cumulative absolute one-step-ahead prediction error for generated multiple target series containing different components. (a)-(f) display results using generated datasets by six different models in equations \eqref{eq:model1-4}, \eqref{eq:model5} and \eqref{eq:model6}. Three other benchmark models (BSTS, ARIMAX and MARIMAX) are also trained and used to make a prediction.}
	\label{fig:Forecast erorr1}
\end{figure*}

\begin{figure*}[h]
	\subfloat[$\rho=0$]{\includegraphics[width = 3in]{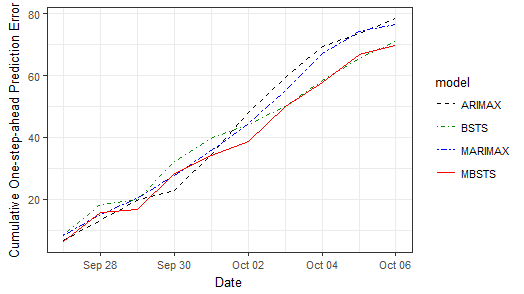}}
	\subfloat[$\rho=0.2$]{\includegraphics[width = 3in]{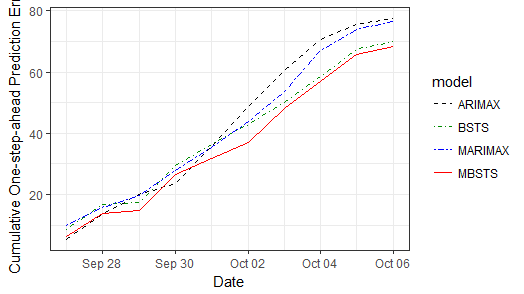}}\\
	\subfloat[$\rho=-0.3$]{\includegraphics[width = 3in]{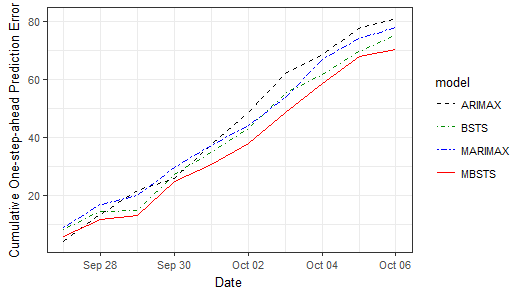}}
	\subfloat[$\rho=0.5$]{\includegraphics[width = 3in]{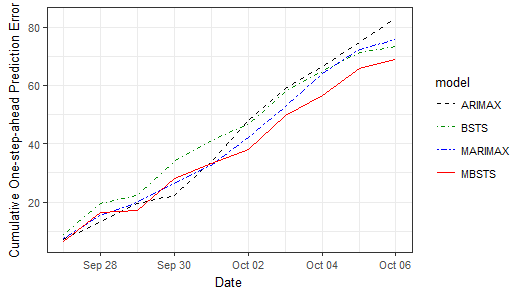}}\\
	\subfloat[$\rho=-0.6$]{\includegraphics[width = 3in]{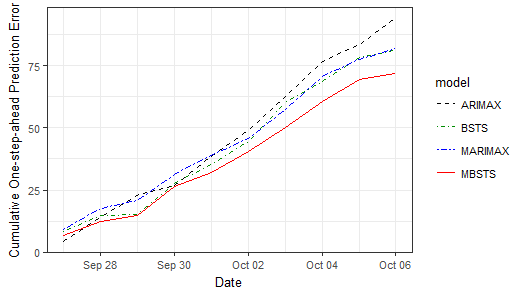}}
	\subfloat[$\rho=0.8$]{\includegraphics[width = 3in]{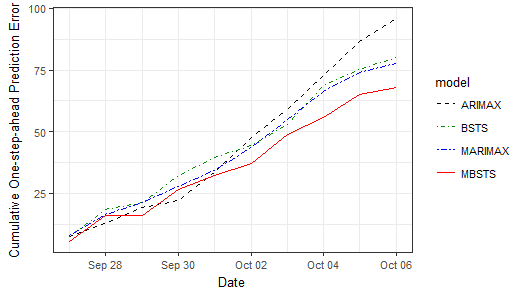}}
	\caption{Cumulative absolute one-step-ahead prediction error for generated multiple target series with different correlations. (a)-(f) display results using generated datasets by equation \eqref{eq:model7} with various correlation coefficients in $\Sigma_t$. Other three benchmark models (BSTS, ARIMAX and MARIMAX) are also trained and used to make a prediction.}
	\label{fig:Forecast erorr2}
\end{figure*}

To evaluate the performance of the MBSTS model, we use three other models: autoregressive integrated moving average model with regression (ARIMAX), multivariate ARIMAX (MARIMAX), and the BSTS model, as benchmark models. We replace ARIMAX and MARIMAX with seasonal ARIMAX (SARIMAX) and multivariate seasonal ARIMAX (MSARIMAX), when seasonality exists.
In this study, applying the growing window approach, all models were trained by the training set and then were used to make a one-step-ahead prediction. More specifically, the univariate BSTS and ARIMAX were trained for each target time series individually, but MBSTS and MARIMAX were applied on the multidimensional series dataset as a whole. Then we compared the performances of the other three models with that of MBSTS in terms of cumulative one-step ahead prediction errors. The prediction error at each step PE$_t$ is defined by summing up the absolute values of the differences between the true values and their own predicted values over all target time series, i.e. $\sum_{i=1}^{m}{|y_t^{(i)}-\hat{y}_t^{(i)}|}$. Figure \ref{fig:Forecast erorr1} and Figure \ref{fig:Forecast erorr2} are generated to demonstrate our model's comparison performance under the influence of complexity in different kinds and numbers of multiple target time series, and under various correlations $(\rho=0, 0.2,-0.3, 0.5, -0.6, 0.8)$, respectively. 

Figure \ref{fig:Forecast erorr1} shows cumulative one-step-ahead prediction errors of six time series models, which were trained using a set of datasets with each containing one thousand observations generated by equations \eqref{eq:model1-4}, \eqref{eq:model5} and \eqref{eq:model6}. 
We can see that the MBSTS model does not show an obvious advantage in the first two plots, since the generated target time series have only a local trend or a linear trend.  
However, the MBSTS model beats other benchmark models in plot $3$ and plot $4$, where the target series contain seasonality or cycle components. Clearly, the BSTS or MBSTS model has a strong ability to capture seasonality and cycle embodied in the series. The performance evaluations in plot $5$ (three target time series) and plot $6$ (four target time series) demonstrate the forecast advantage of our MBSTS model over other benchmark models, even with an increased number of target series.
In general, the multivariate models outperform their corresponding univariate ones due to the influence of correlations among multiple target time series. Moreover, BSTS is better than ARIMAX, and MBSTS outperforms all other models, thanks to the Bayesian model averaging and time series structure of target series.

Figure \ref{fig:Forecast erorr2} provides a clear picture of an impressive fact: the higher correlation among multiple target time series, the better performance of the MBSTS model over other models. Generally, the MBSTS model outperforms the traditional ARIMAX or MARIMAX model for the reason that averaging algorithm helps hedge against selecting the ``wrong" set of predictors in prediction steps. The gaps of cumulative prediction errors between models in a multivariate version and their univariate counterparts increase as multiple target time series have stronger correlations. Therefore, it is better to model multiple target time series as a whole by MBSTS rather than model them individually by BSTS, especially when strong correlations appear in the multiple target time series, as illustrated in Figure \ref{fig:Forecast erorr2}


\section{Application to Empirical Data}
Predicting stock prices (for example, of a group of leading companies) is extremely important to Wall Street practitioners for investment and/or risk management purposes.
In the following, we forecast the future values of stock portfolio return using the proposed MBSTS model and compare its performance with three other benchmark models: BSTS, ARIMAX and MARIMAX.
In this section, we analyze the data of Bank of America (BOA), Capital One Financial Corporation (COF), J.P. Morgan (JPM) and Wells Fargo (WFC). The daily data sample is from 11/27/2006 to 11/03/2017 and obtained from Google Finance.

\subsection{Target Time Series}
We perceive the stock as worthwhile in terms of trading when its future price is predicted to vary more than $p\%$ of its current price. In this context, we forecast the trend of stock movements in the next $k(=5)$ transaction days, which is especially helpful when liquidation risk is in consideration given a sign of sale, and useful to avoid a large amount purchase driving up the stock prices given a sign of buying.  In this study, we provide daily predictions sequentially of the overall price dynamics in the next $k$ transaction days.

\begin{figure*}[h]
	\centering
	\begin{tabular}{c}
		\includegraphics[width=1\linewidth]{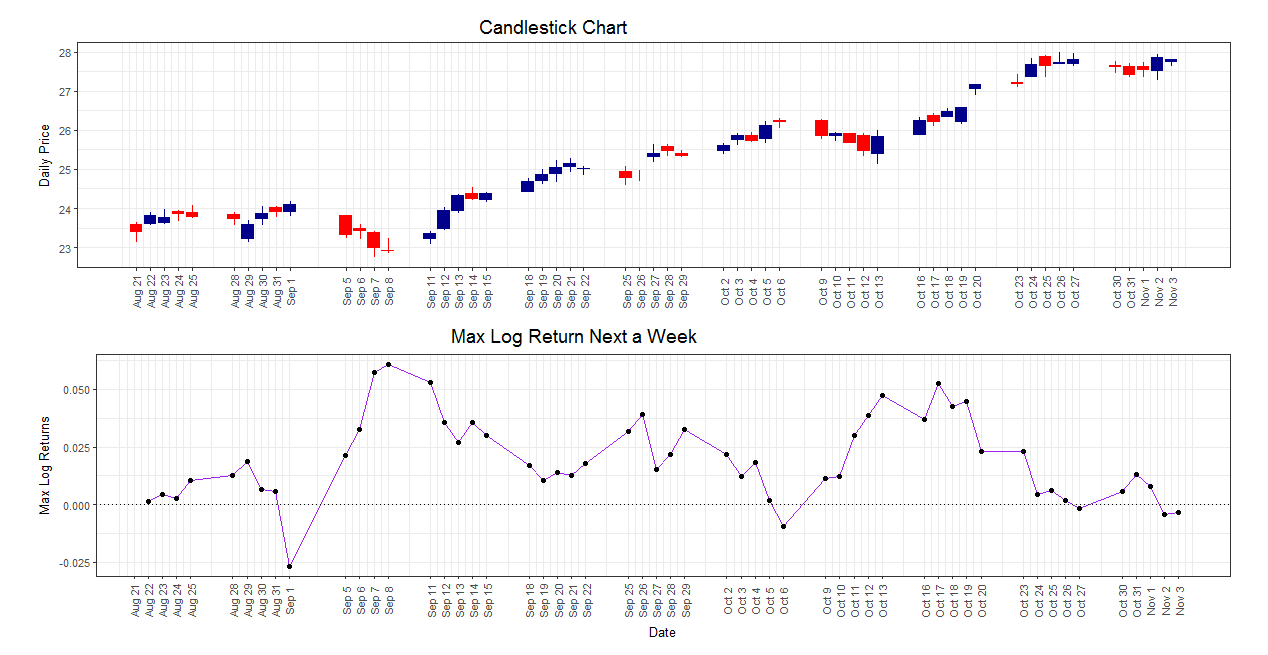}
	\end{tabular}
	\caption{The candlestick chart and max log return. The top panel displays a candlestick chart of BOA from Aug $21$st to Nov $3$rd, containing information such as open and closing quotes. The bottom panel shows corresponding max log returns over the next five transaction days, which is the target time series.}
	\label{fig:candle_stick}
\end{figure*}

Following \cite{torgo2011data}, we approximate the daily average price as:
$
\bar{P}_t =(C_t+H_t+L_t)/3,
$
where $C_t$, $H_t$ and $L_t$ are the close, high and low quotes for day $t$ respectively. However, instead of using the arithmetic returns, we are interested in the log return $V_t$ defined as $
V_t = \{\log(\bar{P}_{t+j}/C_t)\}_{j=1}^k.
$
We consider the indicator variable $
y_t = \max\{v\in V_t\},
$ the maximum value of log returns over the next $k$ transaction days.
A high positive value of $y_t$ means that there is at least one future daily price that is much higher than today's close quote, indicating potential opportunities to issue a buy order, as we predict the prices will rise. A trivial value of $y_t$ around zero can be seen as the sign of no action that should be taken at this moment. In this study, we calculated $y_t $ for four leading companies in the financial industry (BOA, COF, JPM and WFC), whose stock prices are affected by economic activities. Visualization of a part of the daily prices time series and their corresponding $y_t$ indicators for BOA can be seen in Figure \ref{fig:candle_stick}.

\subsection{Predictors} To better capture market information and different properties of the stock price time series and to facilitate the forecasting task, we use the following fundamental and technical predictors.

\paragraph{Fundamental Part}
Fundamental analysis claims that markets may incorrectly price a security in the short run but will eventually correct it. Profits can be achieved by purchasing the undervalued security and then waiting for the market to recognize its ``mistake" and bounce back to the fundamental value. Since macroeconomy has a significant effect on the financial market, economical analysis plays an important role in fundamental analysis in giving a precise stock return prediction.

For economic analysis, we know that it is difficult to collect important economic indicators on a daily basis. However, starting from the year 2004, Google has been collecting the daily volume of searches related to various aspects of macroeconomics. This database is publicly available as ``Google Domestic Trends". In a recent study, \cite{preis2013quantifying} showed correlations between Google domestic trends and the equity market. In this study, we use the Google domestic trends data as a representation of the public interest in various macroeconomic factors, and  include $27$ domestic trends which are listed in Table \ref{table:domestic_trends} with their abbreviations.

\begin{table}[h!]
	\centering
	\begin{tabular}{||c c c c||}
		\hline
		Trend & Abbr. & Trend & Abbr. \\ [0.5ex]
		\hline\hline
		Advertising \& marketing & advert & Air travel & airtvl\\
		Auto buyers & auto & Auto financing & autoby \\
		Automotive & autofi & Business \& industrial & bizind \\
		Bankruptcy & bnkrpt & Commercial Lending & comlnd \\
		Computers \& electronics & comput & Construction & constr \\
		Credit cards & crcard & Durable goods & durble \\
		Education & educat & Finance \& investing  & invest \\
		Financial planning & finpln & Furniture  & furntr\\
		Insurance & insur & Jobs  & jobs\\
		Luxury goods & luxury & Mobile \& wireless  & mobile\\
		Mortgage & mtge & Real estate  & rlest\\
		Rental & rental & Shopping  & shop\\
		Small business & smallbiz & Travel  &  travel\\
		Unemployment & unempl & &\\ [1ex]
		\hline
	\end{tabular}
	\caption{Google domestic trends}
	\label{table:domestic_trends}
\end{table}

\paragraph{Technical Part}
Technical analysis claims that useful information is already reflected in stock prices. We selected a representative set of technical indicators to capture the volatility, close location value, potential reversal, momentum and trend of each stock. Eight variables are calculated for each company as listed in Table \ref{table:eight_variables}:
\begin{table}[h!]
	\centering
	\begin{tabular}{||c c||}
		\hline
		Variable & Abbr.   \\ [0.5ex]
		\hline\hline
		Chaikin volatility & ChaVol  \\
		Yang and Zhang Volatility historical estimator& Vol \\
		Arms' Ease of Movement Value & EMV \\
		Moving Average Convergence/Divergence & MACD \\
		Money Flow Index & MFI  \\
		Aroon Indicator & AROON \\
		Parabolic Stop-and-Reverse & SAR \\
		Close Location Value & CLV \\[1ex]
		\hline
	\end{tabular}
	\caption{Stock Technical Predictors}
	\label{table:eight_variables}
\end{table}

\begin{itemize}
	\item The ChaVol indicator depicts volatility by calculating the difference between the high and low for each period or trading bar, and measures the difference between two moving averages of a volume weighted accumulation distribution line.
	\item The Vol indicator has the minimum estimation error, and is independent of drift and opening gaps, which can be interpreted as a weighted average of the Rogers and Satchell estimator, the close-open volatility, and the open-close volatility.
	\item The EMV indicator is a momentum indicator developed by Richard W. Arms, Jr., which takes into account both volume and price changes to quantify the ease (or difficulty) of price movements.
	\item The MACD indicator is a trading indicator used in stock prices' technical analysis, created by Gerald Appel in the late 1970s, supposed to reveal changes in the strength, direction, momentum and duration of a trend in a stock's price.
	\item The MFI indicator is a ratio of positive and negative money flow over time and starts with the typical price for each period. It is an oscillator that uses both price and volume to measure buying and selling pressure, created by Gene Quong and Avrum Soudack.
	\item The AROON indicator is a technical indicator used to identify trends in an underlying security and the likelihood that the trends will reverse, including ``Aroon up" (resp. ``Aroon down")  for measurement of the strength of the uptrend (resp. downtrend), and reports the time it takes for the price to reach the highest and lowest points over a given time period.
	\item The SAR indicator is a method proposed by J. Welles Wilder, Jr., to find potential reversals in the market price direction of traded goods such as securities.
	\item The CLV indicator is used to measure the closes quote relative to the day's high and low, which varies in range between $-1$ and $+1$.
\end{itemize}

\subsection{Training Result}
It is worth noting that all predictors do not show obvious trends and most of them are stationary in the sense that their unit-root null hypotheses have p-values less than $0.05$ in the augmented Dickey-Fuller test (\cite{said1984testing}). However, some of them indicate seasonal patterns. We can remove seasonal patterns of these predictors by subtracting the estimated seasonal component computed by the STL procedure (\cite{cleveland1990stl}). Then we test our MBSTS model with and without deseasonalizing the predictors.

\begin{figure*}[h]
	\centering
	\begin{tabular}{c}
		\includegraphics[width=1\linewidth]{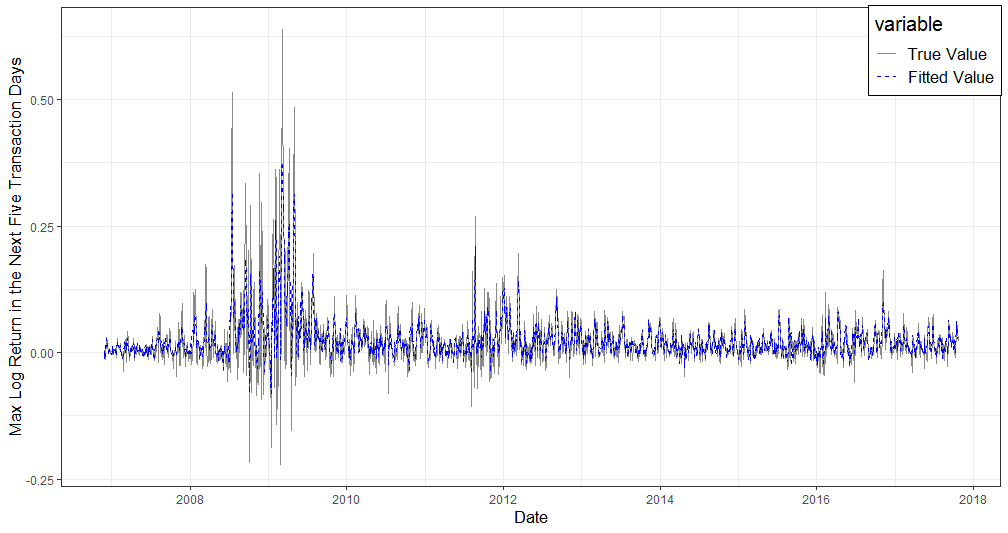}
	\end{tabular}
	\caption{True and fitted values of max log returns from 11/27/2006 to 10/20/2017 (BOA)}
	\label{fig:return1}
\end{figure*}

These eight technical predictors are calculated for each financial institution and then exclusive to others. Domestic Google trends serve as common predictors available to all companies. Based on the forecast output, the model trained without deseasonal predictors performs better than the corresponding one with deseasonal predictors. Therefore, the training results shown in Figure \ref{fig:return1} are from a model with original predictors.

\subsubsection{Decomposition of State Components}
The business cycle describes the fluctuations in economic activities that an economy experiences over a period of time. It typically involves shifts over time between periods of expansions and recessions, which has a great impact on institutions in financial industry, especially investment and commercial banks. We use BOA as an example to illustrate target series and its corresponding state components. Visually checking the time series of max log returns over the next five transaction days in Figure \ref{fig:return1}, we see strong fluctuations during 2008-2009, which is right after the outbreak of the subprime mortgage crisis. There is also an obvious subsequent strong variation during $2012$. Therefore, in order to capture recurrent economic shocks, it is necessary to incorporate the cyclical component in our model. In fact, applying the trend-cycle model can capture both short-term and long-term movements of the series.

\begin{figure*}[h]
	\centering
	\subfloat[Trend Component]{\includegraphics[width = 6.0in]{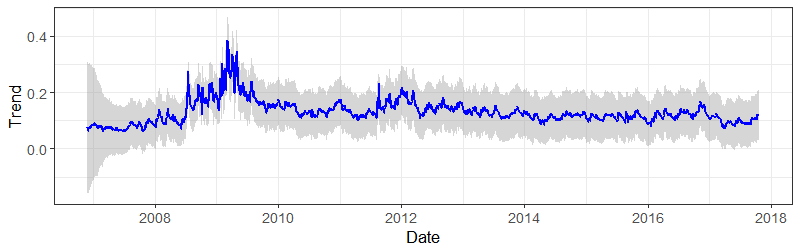}}\\
	\subfloat[Cyclical Component]{\includegraphics[width = 6.0in]{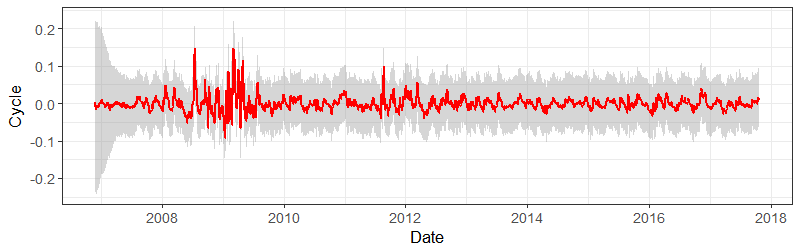}}\\
	\subfloat[Regression Component]{\includegraphics[width = 6.0in]{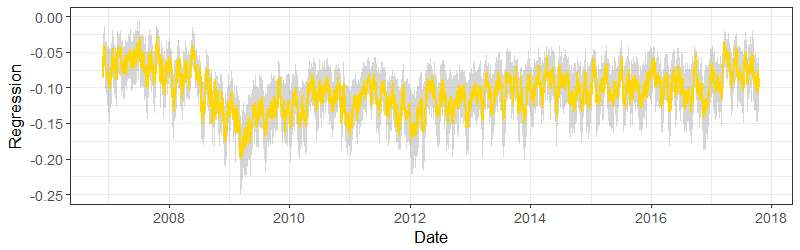}}
	\caption{Contributions of state components to max log return (BOA). The fitted target series is decomposed into three state components: (a) trend (local level in this example) component, (b) cycle component and (c) regression component, 
		with shaded areas indicating the $90\%$ confidence bands based on MCMC draws.}
	\label{fig:stateboa}
\end{figure*}


By spectral analysis, we find the corresponding period equals $274$, which is almost one year of transaction days. Through cross validation, we find the optimal damping factor equals $0.95$ in terms of cumulative one-step prediction errors. Figure \ref{fig:stateboa} shows how much variation in the max log return time series is explained by the trend, cyclical and regression components. The trend component shows the highest peak is around $2009$, and provides a general picture of how the series would evolve in the long run. The comparatively stronger variation between $2009$ and $2012$ is reflected in the cyclical component, which captures the economic shocks that occurred. The fluctuations gradually become stable as the effects of shocks diminish. Both trend and cyclical components handle the series with unequal variances over time. On the contrary, the regression component varies more frequently but with no obvious peaks. It accounts for local movements without the impact of external shocks. In sum, decomposing the target time series into three components provides us enough information on how each component contributes in explaining variations.

\subsubsection{Feature Selection}

Thanks to the spike and slab regression, one advantage of the MBSTS model is that feature selection and model training can be done simultaneously, which prevents overfitting and avoids redundant or spurious predictors. That is, the MBSTS model is flexible in that it selects a different set of predictors for each target time series during the MCMC iterations. Moreover, we can set a different model size for each target time series by assigning appropriate values to the prior inclusion probabilities $\{\pi_{ij}\}$. The empirical posterior inclusion probability, as a useful indicator of the importance of one specific predictor, is the proportion of number of times that the predictor is selected to the total count of MCMC iterations. A higher inclusion probability indicates more variation of target time series can be explained by that predictor, whose chance of being selected depends on equation \eqref{eq:49}.

\begin{figure*}[h]
	\subfloat[Bank of America Corp.]{\includegraphics[width = 3in]{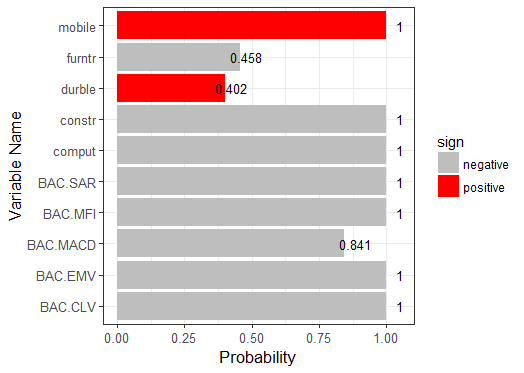}}
	\subfloat[Capital One Financial Corp.]{\includegraphics[width = 3in]{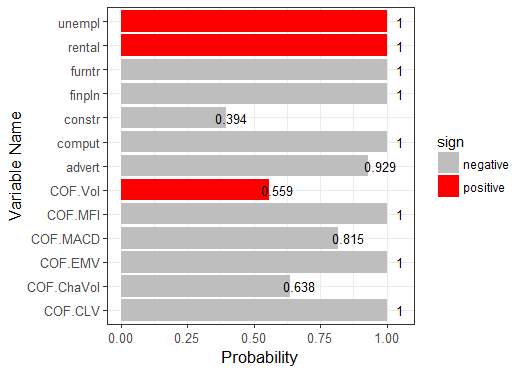}}\\
	\subfloat[JPMorgan Chase $\&$ Co.]{\includegraphics[width = 3in]{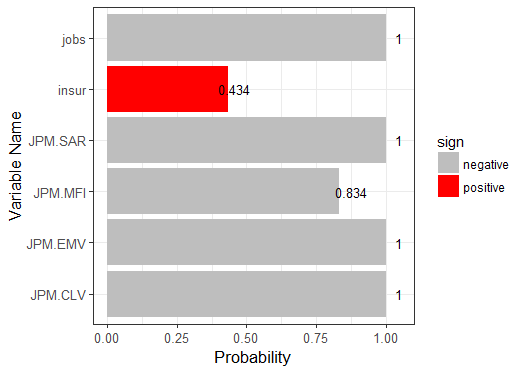}}
	\subfloat[Wells Fargo $\&$ Co.]{\includegraphics[width = 3in]{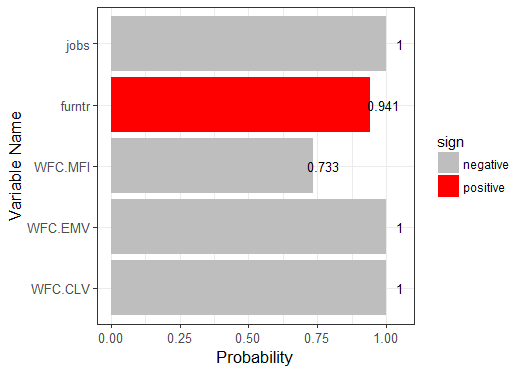}}
	\caption{Empirical posterior inclusion probabilities for the most likely predictors of max log return. (a), (b), (c) and (d) display important predictors for BOA, COF, JPM and WFC respectively. Each bars is colored (red or gray) according to the sign (positive or negative) of the estimated value of the corresponding regression coefficient.}
	\label{fig:prob}
\end{figure*}

Figure \ref{fig:prob} displays the predictors whose empirical posterior inclusion probabilities are greater than $0.2$ for four companies. For the predictors with empirical inclusion probabilities equal to one, we can see that
Bank of America has seven, Capital One Financial Corporation has eight, J.P. Morgan has four, and Wells Fargo has three.
That is, the sets of predictors are different among these four companies; hence, the expected model size for each company also differs from each other. In general, sparsity was produced by our algorithm, and the size of the resulting model for each company is much less that of the total number of candidate predictors.

No such domestic Google trends contribute significantly to the variations of max log returns for all companies. Different sets of domestic Google trends capture the variations of max log returns of these four companies; more specifically, ``mobile", ``constr" and ``comput" are the most important economic indicators for Bank of America, ``unempl", ``rental", ``furntr", ``finpln" and ``comput" for Capital One Financial Corporation, ``jobs" for J.P. Morgan and Wells Fargo.
Among all the technical predictors, MFI, EMV and CLV were favored by the sampling algorithm
for all companies, indicating the importance of these predictors in explaining the variations of max log returns.

\subsection{Target Series Forecast}
Time series forecasting is challenging, especially when it comes to multivariate target time series. One strength of our model is that it can make predictions for multiple target time series (i.e. max log returns of a stock portfolio) with a great number of contemporaneous predictors. Moreover, the Bayesian paradigm together with the spike-slab regression and MCMC algorithm can further improve prediction accuracy through model averaging technique. Similar to the performance analysis on simulated data, we compared the MBSTS model's performance using real financial market data with three other benchmark models: BSTS, ARIMAX and MARIMAX, measured by cumulative one-step-ahead prediction errors.

\subsubsection{Model Comparison}
\begin{figure*}[h]
	\subfloat[All Predictors Without Deaseasonal]{\includegraphics[width = 3in]{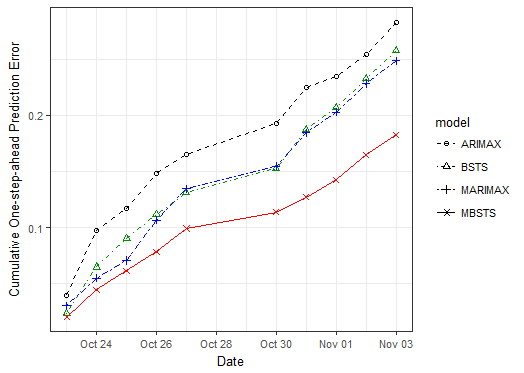}}
	\subfloat[Partial Predictors With Deaseasonal]{\includegraphics[width = 3in]{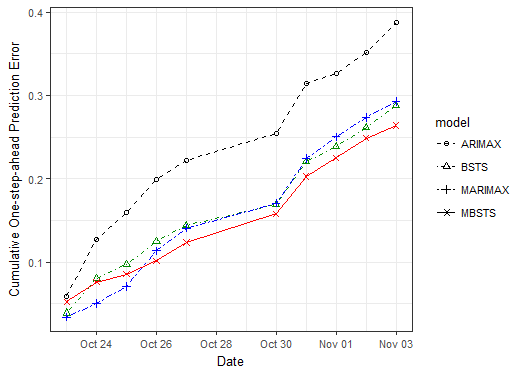}}
	\caption{Performance analysis measured by cumulative one-step-ahead prediction errors: (a) displays the result when all predictors are original; (b) shows the result with some predictors with detected seasonality being deseasonalized. 
		Other three benchmark models (BSTS, ARIMAX and MARIMAX) were also trained to make predictions.}
	\label{fig:forecascompare}
\end{figure*}

Figure \ref{fig:forecascompare} shows the cumulative one-step-ahead prediction errors of these four models without and with deseasonalized predictors, respectively. We can see that the MBSTS model outperforms other benchmark models with
smaller cumulative prediction errors at almost every step in these two cases. We can also see that models with original predictors outperform those using deseasonalized predictors. There are two obvious reasons to explain why the MBSTS model is the best.
Firstly, benefiting from the multivariate setting, it captures the inherent correlations of multiple target time series after subtracting the effects of trend, seasonality and cycle components; these enables MBSTS to outperform the univariate BSTS model that is trained by each target time series individually. Secondly, 
Bayesian model averaging helps avoid arbitrary selection and sticking to a fixed set of predictors, and the cyclical component can capture dramatic shocks to variations in target time series with diminishing impact, both of which enable our MBSTS model to outperform the MARIMAX model.

\begin{figure*}[h]
	\subfloat[Bank of America Corp.]{\includegraphics[width = 3in]{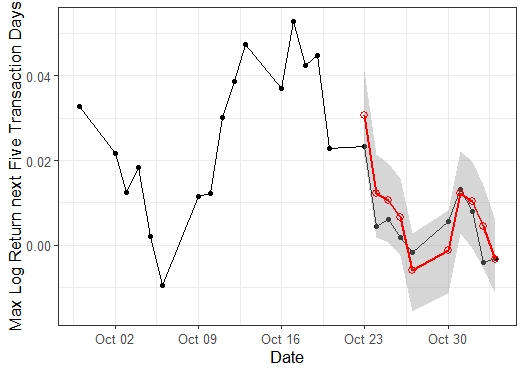}}
	\subfloat[Capital One Financial Corp.]{\includegraphics[width = 3in]{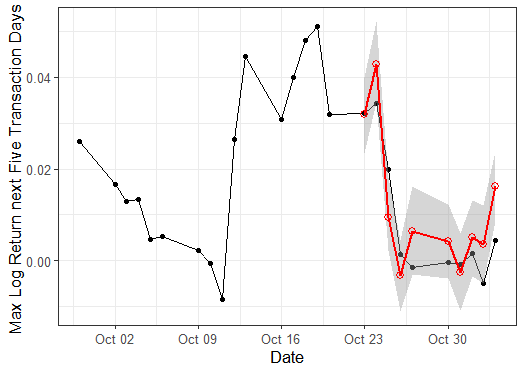}}\\
	\subfloat[JPMorgan Chase $\&$ Co.]{\includegraphics[width = 3in]{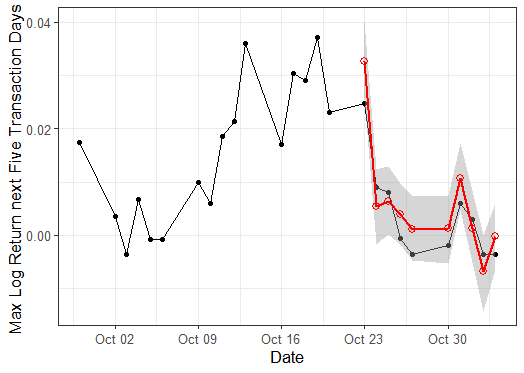}}
	\subfloat[Wells Fargo $\&$ Co.]{\includegraphics[width = 3in]{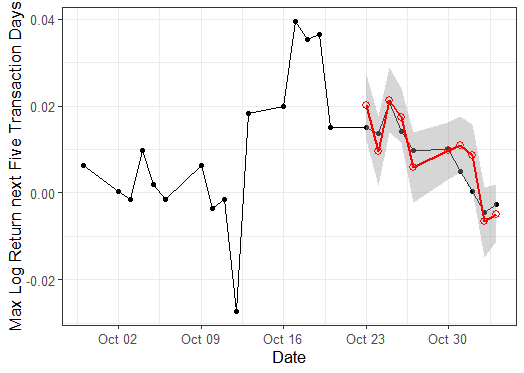}}
	\caption{One-step-ahead predictions of max log returns: (a), (b), (c) and (d) display predicted and true max log return values for BOA, COF, JPM and WFC, respectively. Black lines with dots represent the true values, while red line with dots indicate predicted values. The gray shaded areas are $40\%$ prediction bands.}
	\label{fig:onestepprediction}
\end{figure*}

\subsubsection{Trading Strategy}

In finance, a trading strategy is a set of objective rules defining the conditions that must be met for a trade entry or exit action. Thanks to the strong prediction power, our model can provide supplemental guidelines to trading, given the current information of domestic Google trends and technical indexes. In other words, security strategists can decide when and how to trade based on the predictions from the MBSTS model.

Figure \ref{fig:onestepprediction} shows one-step-ahead predictions by the MBSTS model for these four companies over two weeks. The shaded areas are the $40\%$ prediction intervals generated by draws from the posterior distribution of $\hat{y}$. All true values are covered by the prediction intervals. The predicted value of max log return can be used as an indicator of whether to trade a stock or not. For example, if the lower bound of the predicted max log return is a large positive number, it is a strong signal that future prices will go substantially above the closing price of that day, thus buying this stock that day should be seriously considered.
When the predicted value is positive but not large enough to cover transaction cost, it is a weak buying signal and a second thought should be given before making a decision. Selling or shorting the stock is suggested if the predicted max log return in the next five transaction days, is negative.

\section{Conclusion}

In this paper, we have proposed a Multivariate Bayesian Structural Time Series (MBSTS) model for dealing with multiple target time series (e.g. max log returns of a stock portfolio), which helps in feature selection and forecasting in the presence of related external information. We evaluated the forecast performance of our model using both simulated and empirical data, and found that the MBSTS model outperforms three other benchmark models: BSTS, ARIMAX and MARIMAX.
This superior performance can be attributed mainly to the following three reasons. Firstly, the MBSTS model derives its strength in forecasting from the fact that it incorporates information about other variables, rather than merely historical values of its own. Secondly, the Bayesian paradigm and the MCMC algorithm can perform variable selection at the same time as model training and thus prevent overfitting even if some spurious predictors are added into the candidate pool. Thirdly, the MBSTS model benefits from taking correlations among multiple target time series into account, which helps boost the forecasting power.
Therefore, this model, as expected, is able to provide more accurate forecasts than the univariate BSTS model and the traditional time series models such as ARIMA or MARIMA, when multiple target time series need to be modeled.

The excellent performance of the MBSTS model comes with high computation requirements in the MCMC iterations. Clearly, one would also not expect this model to show significant advantages over the univariate BSTS model, when multiple target series are independent of each other. But some preliminary exploratory analysis as well as professional insight would help to tell whether
correlations in multiple target time series are strong enough in specific cases.  Two open questions that are currently under investigation include: 
whether and how prior information such as model size and estimated coefficients can improve estimation accuracy and forecasting performance; the other is how to adjust this model to satisfy the need of analysis of non-Gaussian observations. Overall, it is fair to conclude that the MBSTS model offers practitioners a very good option to model or forecast multiple correlated target time series with a pool of available predictors.

\section*{Acknowledgements}
We would like to thank the journal editor and the anonymous reviewers who
provided us with many constructive and helpful comments.
\vskip 0.2in
\bibliography{mbsts}

\end{document}